\documentclass[10pt,twocolumn,twoside] {IEEEtran}
\def\comment#1{}
\usepackage{graphicx,subfigure,subfig,epsfig,amsfonts,amsmath,amssymb,lscape,color,url}
\usepackage[noadjust]{cite}
\def\comment#1{}

\newcommand{\Dmat}{{\bf D}}

\newcommand{\Imat}{{\bf I}}

\newcommand{\Xmat}{{\bf X}}

\newcommand{\dv}{\boldsymbol{d}}

\newcommand{\Wv}{\boldsymbol{W}}
\newcommand{\xv}{\boldsymbol{x}}
\newcommand{\yv}{\boldsymbol{y}}

\newcommand{\Psimat}{\boldsymbol{\Psi}}

\newcommand{\epsilonv}{\boldsymbol{\epsilon}}

\begin{document}
\setlength{\parskip}{.02in}

\title{Compressive Hyperspectral Imaging\\ with Side Information}

\author{
\authorblockN{Xin Yuan, Tsung-Han Tsai, Ruoyu Zhu, Patrick Llull, David Brady, and Lawrence Carin} \\
\authorblockA{Department of Electrical and Computer Engineering, 
Duke University \\
Durham, NC 27708-0291 USA }
}

\maketitle

\begin{abstract}
A blind compressive sensing algorithm is proposed to reconstruct hyperspectral images from spectrally-compressed measurements.
The wavelength-dependent data are coded and then superposed, mapping the three-dimensional hyperspectral datacube to a two-dimensional image.
The inversion algorithm learns a dictionary {\em in situ} from the measurements via {\em global-local shrinkage} priors.
By using RGB images as {\em side information} of the compressive sensing system, the proposed approach is extended to learn a coupled dictionary from the joint dataset of the compressed measurements and the corresponding RGB images, to improve
reconstruction quality.
A prototype camera is built using a liquid-crystal-on-silicon modulator.
Experimental reconstructions of hyperspectral datacubes from both simulated and real compressed measurements demonstrate the efficacy of the proposed inversion algorithm, the feasibility of the camera and the benefit of side information.
\end{abstract}

\begin{IEEEkeywords}
Compressive sensing,
hyperspectral image,
side information,
Bayesian shrinkage,
dictionary learning,
blind compressive sensing,
computational photography,
coded aperture snapshot spectral imaging (CASSI),
spatial light modulation.
\end{IEEEkeywords}

\maketitle
\section{Introduction}
Hyperspectral imaging techniques have been widely applied in various fields such as astronomy \cite{Hege03}, remote sensing \cite{Goetz85}, and biomedical imaging \cite{Sorg05}. Unlike ordinary imaging systems, hyperspectral imagers capture a three-dimensional (3D) datacube, {\em i.e.}, a 2D array of vectors that contain the spectral information at each spatial location. Inspired by compressive sensing (CS)~\cite{Donoho06ITT,Candes06ITT}, researchers have adopted joint sensing and reconstruction paradigms that measure a subset of the 3D spectral datacube~\cite{Gehm07,Wagadarikar08CASSI,Wagadarikar09CASSI,Li12TIP} 
and utilize CS inversion algorithms to retrieve a 3D estimate of the underlying hyperspectral images. This sensing paradigm follows the traditional benefits of CS - reduced data rates and system complexity at the expense of computational algorithmic development and postprocessing.

\begin{figure}[ht!]
       \centering
       \includegraphics[scale = 0.6]{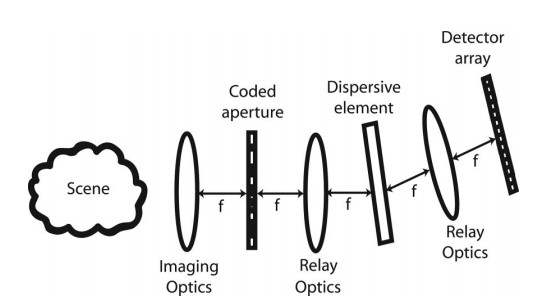}
       \caption{Schematic of CASSI~\cite{Wagadarikar09CASSI}. The imaging optics image the
       scene onto the coded aperture. The relay optics relay the image
       from the plane of the coded aperture to the detector through the
       dispersive element (figure from~\cite{Wagadarikar09CASSI}).}
      \label{Fig:CASSI_old}
\end{figure}

The coded aperture snapshot spectral imaging (CASSI)~\cite{Wagadarikar08CASSI,Wagadarikar09CASSI} systems are examples of CS hyperspectral imaging systems.
The CASSI paradigm encodes each of the datacube's spectral channels with a unique 2D pattern, which is the underlying operating principle behind code division multiple access (CDMA).  CASSI systems form an image onto a coded aperture placed at an intermediate image plane to spatially modulate the datacube with high-frequency patterns (Fig.~\ref{Fig:CASSI_old}). A disperser placed in a pupil plane behind the coded aperture spectrally shifts the coded image, effectively granting each spectral channel onto its own unique coding pattern when multiplexed onto a monochrome sensor.  
Unlike other hyperspectral imaging techniques~\cite{Green98,Herrala94,Morris94}, CASSI can obtain a discrete 3D estimate of the target spectral datacube from as little as a single 2D measurement.  Such operation renders the system's forward model highly underdetermined; inversion requires use of CS  algorithms~\cite{Bioucas-Dias2007TwIST,Figueiredo07GPSR,Yin08bregman,Yuan14CVPR}. Compressive hyperspectral imaging requires the signal under estimation to be sparse in a basis that is incoherent with the system sensing matrix~\cite{Wagadarikar08CASSI}. The reconstruction is accomplished using optimization algorithms, such as gradient projection for sparse
reconstruction (GPSR)~\cite{Wagadarikar08CASSI} or two-step iterative shrinkage/thresholding (TwIST)~\cite{Kittle10AO}. GPSR assumes
sparsity of the entire datacube in a fixed (wavelet) basis, while TwIST is based on a piecewise-constant spatial
intensity model (when the total variation norm is used) for hyperspectral images.

Distinct from the above optimization algorithms, blind CS algorithms~\cite{Rajwade13SIAM} have been applied to CASSI systems by learning dictionaries from the measurements.
Blind CS inversion strategies seek to recover 3D patches of the hyperspectral datacube jointly with the shared dictionary (inferred from the measurements). Each patch is a sparse combination of the dictionary atoms. Since the dictionary is {\em unknown} \emph{a priori}, this is called {\em blind} CS~\cite{Gleichman11BCS}.
This blind CS model shares statistical strengths among similar image patches at different spatial locations.  Additionally, this CS approach is task-driven since the learned dictionary is appropriate for different tasks~\cite{Duarte-Carvajalino13SPT}.
In this paper, we propose a new blind CS model that imposes {\em compressibility}~\cite{cevher2009learning}, rather than sparsity, on the recovered dictionary coefficients.  
Specifically, we use {\em global-local shrinkage} priors \cite{Polson10shrinkglobally,Polso12Levy,Yuan14TSP} on the dictionary coefficients for each patch, under a Bayesian dictionary learning framework. This setting drives the reconstructed dictionary coefficient vector to contain many small ({\em i.e.}, close to zero) components while allowing for a few large components, but avoiding explicit sparsity enforcement. This model is feasible for high-dimensional hyperspectral signals since it can extract more information from the limited ({\em in situ} learned) dictionary atoms.

The quality of the reconstructed hyperspectral images relies on the conditioning of the sensing matrix.  More accurate recovery is possible with additional measurements of the same scene \cite{Kittle10AO} or by using digital micromirror (DMD) arrays~\cite{Wu11}; however, these methods increase system cost and energy consumption.
This paper features a blind CS algorithm that employs the RGB image of the target scene (as {\em side information}) and demonstrates substantial improvements on reconstruction quality with a single CASSI measurement.
Particularly, the proposed algorithm learns a joint dictionary from a single CASSI measurement and the corresponding RGB image (of the same scene) and then reconstructs the hyperspectral datacube. Since the RGB image is strongly correlated with the hyperspectral images, this joint model will dramatically aid the dictionary learning algorithm, thereby improving the reconstruction.

Furthermore, we propose a new camera using spatial-light modulation (SLM)~\cite{Lazarev12} for an active coding compressive spectral imager, to jointly modulate the spatial and spectral information of the scene on the intermediate image plane. This technology differs from CASSI and DMD-modulated CS imagers in that no dispersive element is required for multiplexing the spectral channels.

The reminder of the paper is organized as follows. Section~\ref{Sec:Hardware} reviews the mathematical model of CASSI and introduces the new camera. Section~\ref{Sec:Algorithm} develops the new blind CS algorithm to reconstruct hyperspectral images. Experimental results are presented in Section~\ref{Sec:Results}, and Section~\ref{Sec:Col} summarizes the paper.
\section{Hardware \label{Sec:Hardware}}
In this section, we first review the CASSI imaging process and then introduce the new camera. Finally, a shared mathematical model for both cameras is proposed.
\subsection{Review of CASSI}
A common CASSI design~\cite{Wagadarikar08CASSI,Wagadarikar09CASSI,Kittle10AO} in Fig.~\ref{Fig:CASSI_old} adapts the available hardware to encode and multiplex the 3D spatiospectral information $f(x,y;\lambda)$ onto a 2D detector.
This modulation process is based on physically shifting the hyperspectral datacube multiplied by the binary coding pattern $T(x,y)$ via chromatic dispersion.
The coding pattern consists of an array of square, binary (100\% or 0\% transmission) apertures, which provide full or zero photon transmission.
The camera's disperser (in traditional hyperspectral CS~\cite{Gehm07,Gehm08}, this is a prism or diffraction grating) laterally displaces the image as a function of wavelength $d(\lambda-{\lambda}_{c})$ of the coded aperture pattern, where $d(\lambda)$ is the lateral wavelength-dependent shift and ${\lambda}_{c}$ represents the disperser's central wavelength.
As previously mentioned, this coding process can be considered a form of CDMA, whereby each channel is modulated by an independent coding pattern on the detector plane.
The detector integrates the spectrally-shifted planes along the spectral axis. Information can be recovered by separating channels based on their projected coding patterns.
This multiplexing process can be represented as the following mathematical forward model:
\begin{equation}
g(x,y)=\int T(x+d(\lambda-{\lambda}_{c},y))f(x+d(\lambda-{\lambda}_{c},y;\lambda))\mathrm{d}\lambda,
\end{equation}
where $g(x,y)$ is the continuous form of the detector measurement, representing the sum of dispersed coded images.
Since the detector array has been spatially pixelated by the detector pixel pitch $\Delta$, the discreatized detector measurement becomes: 
\begin{equation}
g_{m,n}=\iint g(x,y) {\rm rect}(\frac{x}{\Delta}-m, \frac{y}{\Delta}-n)\mathrm{d}x\mathrm{d}y.
\end{equation}
Finally, the discrete measurement for each pixel can be illustrated as:
\begin{equation}
g_{m,n}=\sum_{j} T_{(m+j-1),n} f_{(m+j-1),n,j}+\epsilon_{m,n}, \label{Eq:CASSI}
\end{equation}
where $T_{m,n}$ is the spatially encoding pattern, $f_{m,n,j}$ represents the discretized spectral density of $f(x,y;\lambda)$, and $\epsilon_{m,n}$ is the noise.
\subsection{A New Camera: SLM-CASSI}
Here we use an SLM to encode the 3D spatial-spectral information on a 2D gray-scale detector - a similar CDMA strategy to CASSI for snapshot hyperspectral imaging.  An SLM is an array of micro cells placed on a reflective layer; each cell has a nematic liquid crystal that adds pixel-level phase to the incident light as a function of voltage, thereby changing the polarization state on a per-pixel basis \cite{Lazarev12}. 
Each layer of the LC can be considered as a thin optical birefringence material; the orientation and the relative refractive index difference between the fast and slow axes determines the effective birefringence.
Since most birefringent phase modulators are nominally sensitive to wavelength, this element can assign wavelength-dependent transmission patterns to multiplex every spectral channels for the compressive measurement. The hyperspectral slices can be separated from the coded data via CS inversion algorithms.
\subsubsection{Experimental Setup}
\begin{figure}[h]
	\center
	\includegraphics[width=3.5in]{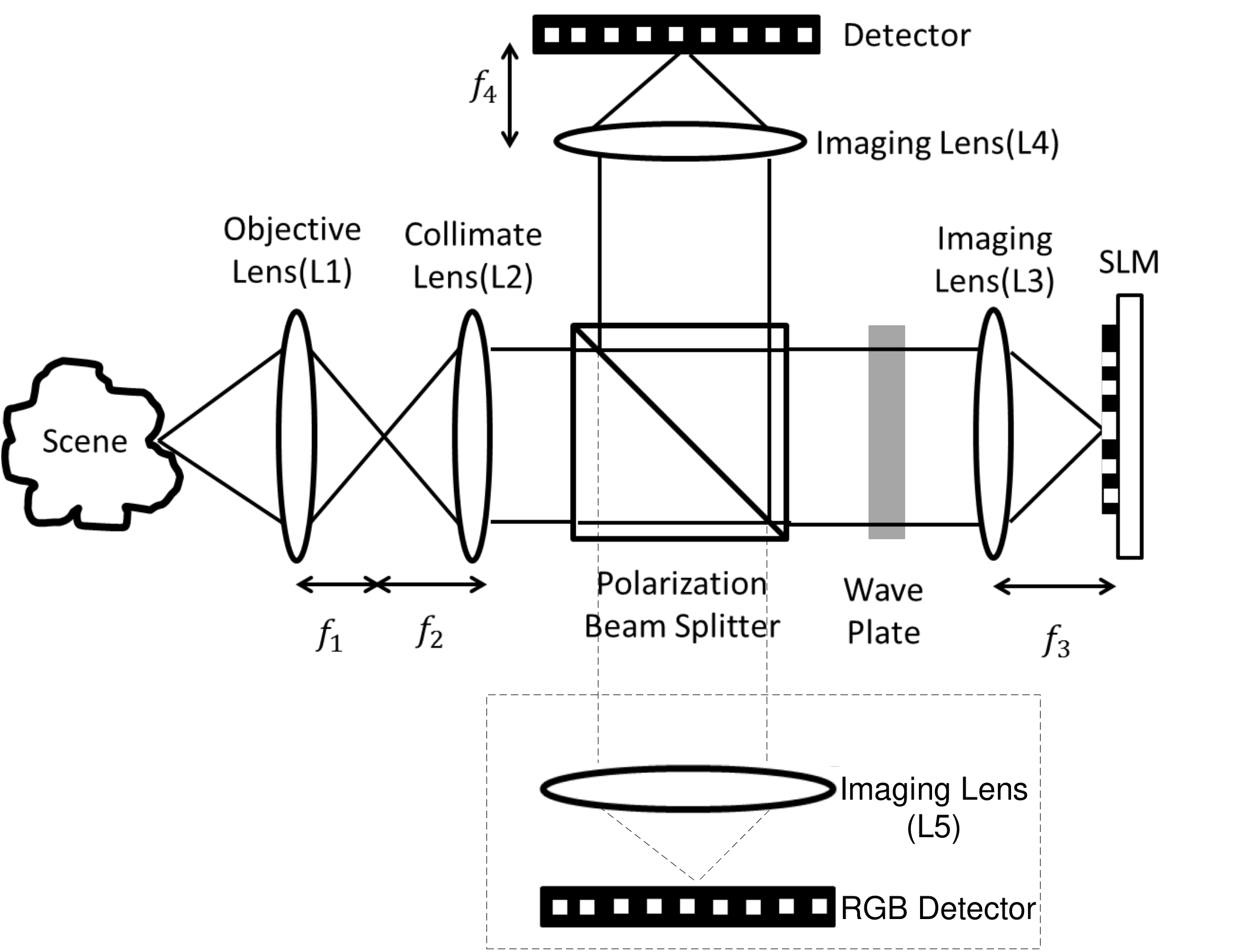}
	\caption{The system schematic of CASSI with SLM. The parts in the dashed box represent ongoing work.}
	\label{fig:System}
\end{figure}
\begin{figure}[h]
	\centering
	\includegraphics[width=3.2in]{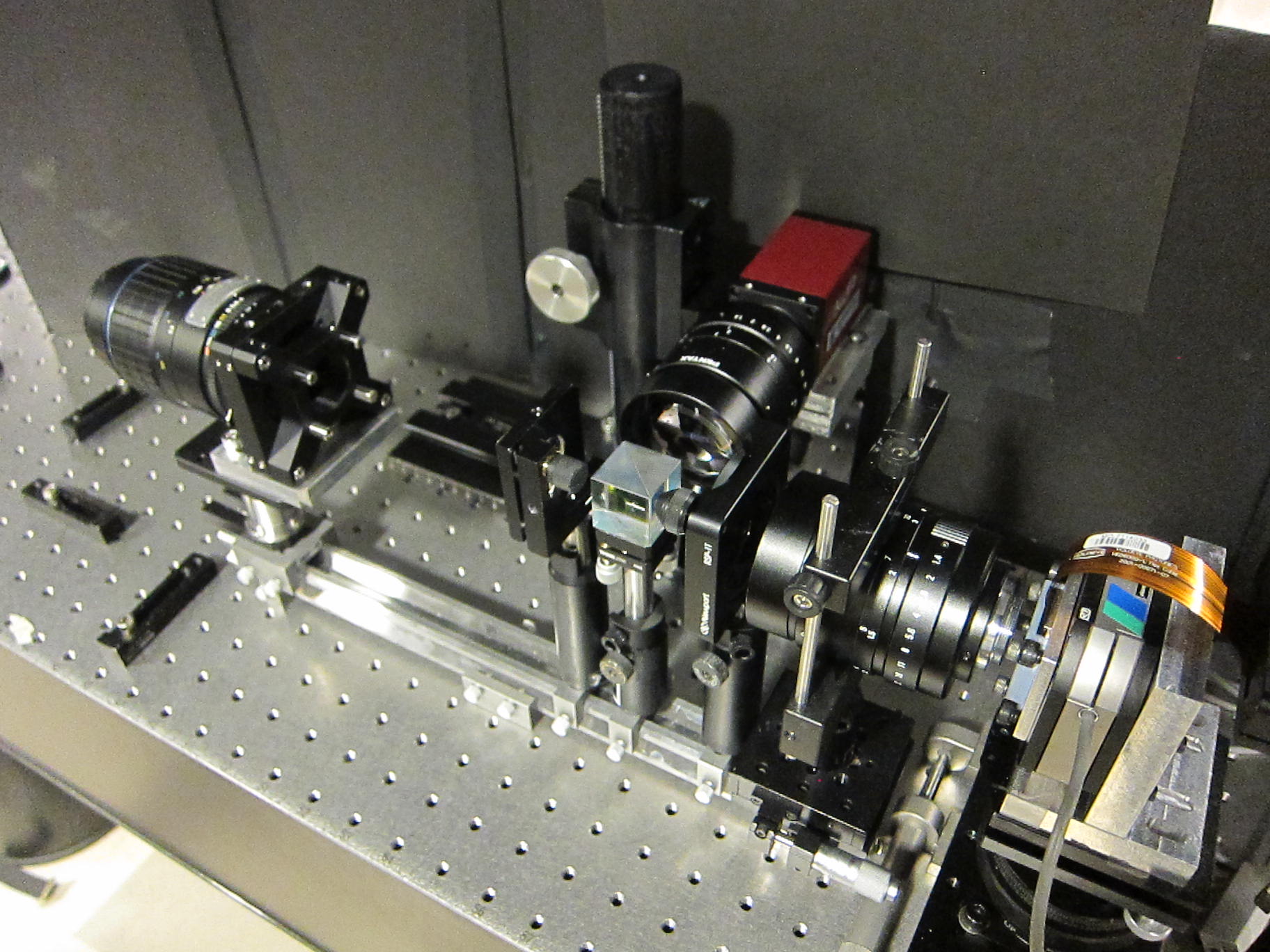}
	\caption{The experimental prototype of the system.}
	\label{fig:Systemsetup}
\end{figure}

A schematic of this SLM-based CASSI system is shown in Fig.~\ref{fig:System}.
An objective lens (L1) images the scene from a remote distance, and then the collimation lens (L2) and imaging lens (L3) relay the image through the polarizing beamsplitter and the achromatic quarter-wave retarder onto the SLM.  The retarder serves to increase the contrast of the SLM polarization array by compensating the extra phase retardation in the liquid crystal device.
The spatiospectral coding function is implemented on the SLM as an 8-bit, pseudo random, pixel-level grayscale phase retardation pattern.
The phase retardation manifests as wavelength-dependent ampitude modulation upon re-entry through the polarizing beamsplitter toward the camera.  Finally, the modulated image is projected by the imaging lens (L4) onto the detector plane and then recorded by the detector.
The following mathematical forward model can be used to describe the multiplexing process of the compressive sampling:
\begin{equation}
g(x,y)=\int T(x,y,\lambda)f(x,y;\lambda)\mathrm{d}\lambda,
\end{equation}
where $T(x,y,\lambda)$ are the wavelength dependent transmission patterns provided by the SLM, which can be calibrated by analyzing its electrically-controlled birefringence.
Since the detector array is spatially pixellated, the $(m,n)^{th}$ detector measurement is given by:
\begin{equation}
g_{m,n}=\sum_{j} T_{m,n,j} f_{m,n,j}+\epsilon_{m,n}, \label{Eq:SLM-CASSI}
\end{equation}
where $T_{m,n,j}$ represents the discretized transmission patterns.
Fig.~\ref{fig:Systemsetup} is a photograph of the experimental setup.
The setup includes a 60 mm objective lens (Jenoptik), a 75 mm achromatic relay lens (Edmound Optics), a polarization beam splitter (Newport), an achromatic quarter wave plate (Newport), a liquid crystal based SLM (Pluto, Holoeye), two 75 mm imaging lenses (Pentax), and a monochrome CCD camera (Pike F-421, AVT) with 2048$\times$2048 pixels that are 7.4 $\mu$m square.
A 450-680nm band pass filter (Badder) is mounted on the objective lens to block unwanted ultra-violet and infrared light (the SLM and camera are optimized for visible-range operation). 
The SLM provides strong modulation and effective multiplexing to fulfill the requirement of compressive sensing. This modulator is based on a reflective Liquid Crystal on Silicon (LCoS) microdisplay technique\cite{Lazarev12}, which has a 1920$\times$1080-pixel active area with an 8 $\mu$m pixel pitch.
\subsubsection{System Calibration}
\begin{figure}[h]
	\center
	\includegraphics[width=3.7in, height = 2.3in]{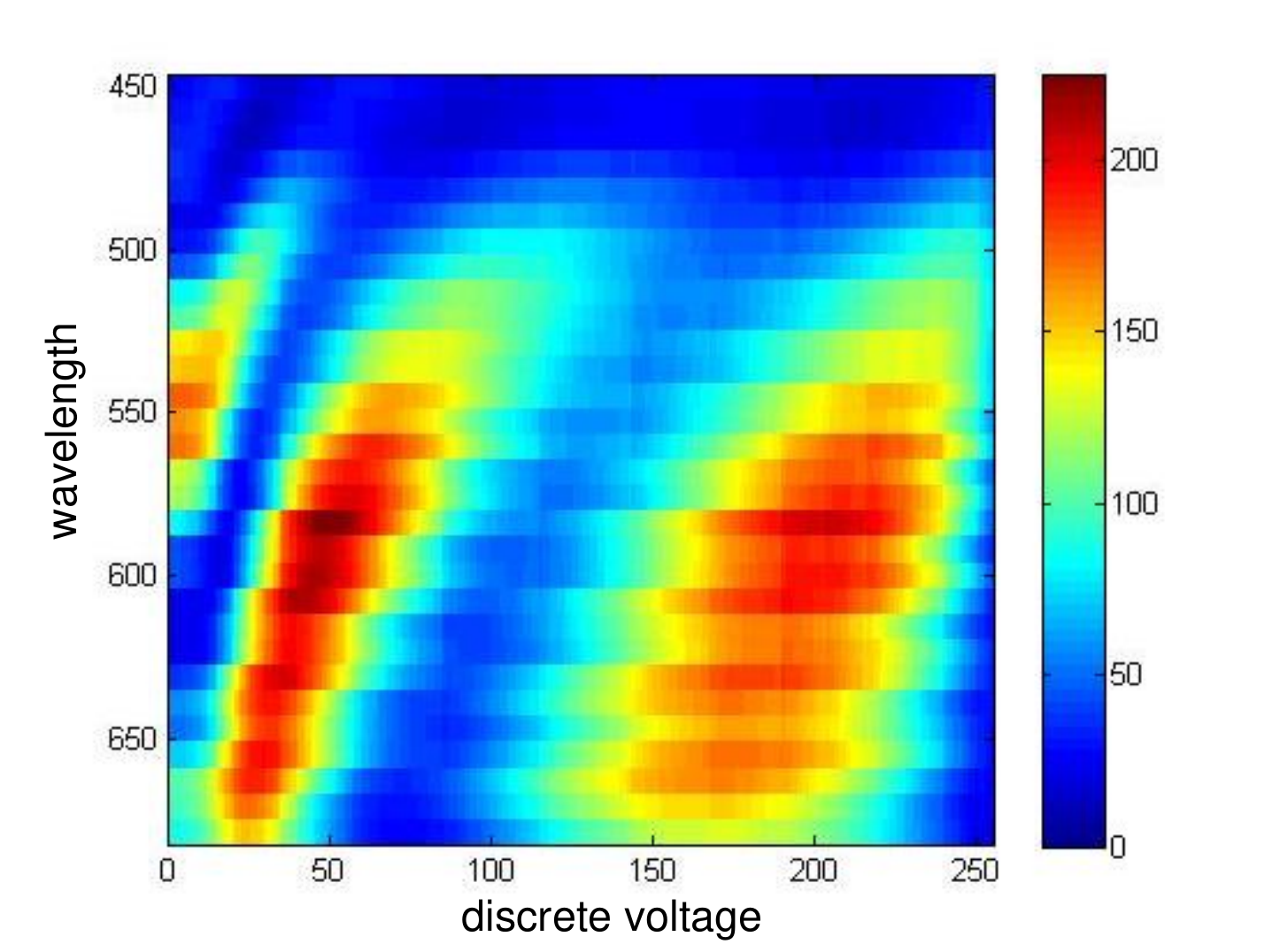}
	\caption{Spectral dependence of SLM amplitude modulation. Given different voltages and incident wavelengths (in nm), the SLM combined with the optics generates corresponding transmission code. Within 450 nm and 680 nm, an 8-bit voltage variation on the SLM generates a gray scale transmission ratio between 0.08 and 0.96. }
	\label{fig:SLMResponse}
\end{figure}
The effect of the 8-bit, pseudo-random voltage pattern on the spectral channels can be estimated by recording the system response generated by the SLM. Fig.~\ref{fig:SLMResponse} records the transmission under the monochromatic illumination and homogeneously-applied voltage on the modulator.
We average the transmission across the center area of the SLM and calculate its relative transmisson.
A potential advantage of this SLM is that we can change the code easily to get different modulations.\footnote{This also opens the door of adaptive compressive sensing~\cite{Duarte-Carvajalino09IPT,Averbuch12SIAM,Carson11SIAM}. However, after plenty experiments, we found the contribution of adaptive sensing is very limited in the CS inversion, partially due to the mismatch between the designed code and the calibration. 
We further found random coding performing well in our camera.}

We assume that the system operator provides one-to-one mapping (1:1 magnification of the SLM pixels onto the detector) between the micro-display and the detector array. 
Theoretically, the system operator $T(x,y,\lambda)$ can be estimated by using SLM's calibration data and the applied voltage on the SLM. However, one might account the error in the real system projection. For example, optical aberrations and sub-pixel alignment discrepancies between the detector and SLM might break the ideal mapping and image some of the SLM pixels onto several detector pixels. Therefore, part of the transmission code might deviate from the ideal value, which results in an inaccurate system operator. Since the system operator dominates the quality of the object estimation in this inverse problem, a careful calibration of the $T(x,y,\lambda)$ is required.

A better representation of the system response can be acquired by recording the transmission pattern illuminated by each spectral band. The revised system operator $T(x,y,\lambda)$ includes all the possible coding patterns generated by the SLM, which can contribute to the data reconstruction. Here we combine the tungsten light source (LSH-T250, Horiba) with a monochromator (iHR320, Horiba) to quantize the spectral dimension into bands of finite width. Each band has a 7.5-8 nm full width at half maximum (FWHM). The scene's spectral irradiance is recorded by a fiber optics spectrometer (USB2000, Ocean Optics); these values are taken as ground truth in the experiments presented later in the paper.
The number of spectral channels and their central wavelengths are determined by the grating period (1800 periods/mm) and the optical path length of the monochromator. To better represent the continuous light source, two adjacent spectral bands are separated by the monochromator's FWHM.  At this calibration resolution, the system's spectrally-sensitive bandwidth (450-680nm) has been discretized into $30$ spectral channels.   Importantly, the spectral resolution of the reconstructed data is determined by the monochromator's resolving power during calibration; smaller FWHM values result in larger numbers of calibrated and reconstructed spectral channels.  Compared to the spectral imagers that are reliant upon dispersive elements to spatially shear the physical code (in CASSI), this SLM based spectral imager can easily improve the spectral resolution without revising the main camera's optical design.
\subsection{Shared Mathematical Model of CASSI and SLM-CASSI \label{Sec:CASSI}}
Assuming we are measuring a hyperspectral datacube ${\bf X}\in {\mathbb{R}}^{N_x \times N_y \times N_{\lambda}}$, where $N_{\lambda}$ signifies the number of spectral channels of spatial extent $N_x \times N_y$.
We can interpret both CASSI and SLM-CASSI measurements as a summation of the 3D Hadamard product of the datacube with the different modulation codes, ${\bf M} = \sum_{j =1}^{N_{\lambda}} {\bf X}_j \odot {\bf C}_j$.
Here, the ${{\bf C}}_j$ is a shifted version of the same code as in CASSI~\cite{Rajwade13SIAM}; however, each spectral modulation pattern of ${{\bf C}}_j$ depends on the SLM response in SLM-CASSI.
For each pixel, both (\ref{Eq:CASSI}) and (\ref{Eq:SLM-CASSI}) can be represented as:
\begin{equation}
{\bf M}(m,n) = \sum_{j=1}^{N_{\lambda}} {\bf X}_j(m,n) {\bf C}_j(m,n).
\end{equation}
The pixel-wise modulation of CASSI results in feasible patch-based reconstruction algorithms.

Since the SLM-CASSI sensing paradigm reconstructs $N_\lambda$ spectral images from a single measurement ${\bf M}$, the compression ratio of the CASSI and SLM-CASSI systems discussed above is $N_{\lambda}$.
The compressive sampling process in both spectral imaging systems can be represented by the same matrix described in the following section.
\section{Reconstruct Hyperspectral Images with Blind Compressive Sensing \label{Sec:Algorithm}}
In this section, we develop a new blind Bayesian CS model to reconstruct hyperspectral images from CASSI or SLM-CASSI measurements. 
The proposed model is a generalized framework which can also be used in denoising and inpainting problems, among others~\cite{Zhou12TIP}.
\subsection{Dictionary Learning Model \label{Sec:DL}}
We decompose the 3D  hyperspectral datacube into $N$ small patches with size $n_x\times n_y\times N_{\lambda}$.  In vector form, these patches are denoted $\boldsymbol{x}_n \in {\mathbb R}^{P}, \forall n=1,\dots, N$, with $P = n_xn_yN_{\lambda}$.
We seek a dictionary $\Dmat\in {\mathbb R}^{P\times K}$ with $K$ atoms  such that ${\boldsymbol{x}}_n = {\Dmat \boldsymbol{s}_n}$, based on the captured image ${\bf M}$. 
For each patch, we have the corresponding 2D measurement patch $\{{\bf M}_n\}_{n=1}^N\in {\mathbb R}^{n_x\times n_y}$ and the 3D mask (coding) patch $\{{\bf C}_n\}_{n=1}^{N} \in \{0,1\}^{n_x\times n_y \times N_{\lambda}}$. 
Considering the vectorized format of the measurement for $n^{th}$ patch ${\yv}_n \in {\mathbb R}^{n_xn_y}$, and the corresponding mask patches $\{\Wv_n^{(j)}\}_{n=1}^N \in \{0,1\}^{n_xn_y}, \forall j = 1,\dots, N_{\lambda}$ (each $\Wv_n^{(j)}$ is the vector form of $j^{th}$ 2D slice in ${\bf C}_n$), we have the measurement model
\begin{eqnarray}
\yv_n &=& \Psimat_n  \xv_n + \epsilonv_n, \label{Eq:MeaModel}\\
\Psimat_n &=& \left[{\rm diag}\{\Wv_n^{(1)}\}, \dots, {\rm diag}\{\Wv_n^{(N_{\lambda})}\}\right],\\
{\bf Y} &=& [\yv_1, \dots, \yv_N],\quad
{\bf X} = [\xv_1, \dots, \xv_N],
\end{eqnarray}
where $\epsilonv_n$ represents the additive Gaussian noise, and we model it as ${\epsilonv}_n \sim {\cal N}(0, \alpha_0^{-1} {\bf I}_P)$, with $\alpha_0$ denoting the noise precision.\footnote{The extension of our model to non-uniform denoising problem~\cite{Mairal07TIP} is straightforward, {\em i.e.}, by imposing spatially-varying noise models for different patches.} We place a diffuse gamma prior on $\alpha_0$
\begin{eqnarray}
\alpha_0 &\sim &{\rm Ga}(c_0,d_0),
\end{eqnarray}
where $(c_0,d_0)$ are hyperparameters.
Taking account of the dictionary learning model, (\ref{Eq:MeaModel}) can be written as:
\begin{equation}
\yv_n = \Psimat_n \Dmat {\boldsymbol{s}}_n + \epsilonv_n, \label{eq:ywDs}
\end{equation}
where ${\boldsymbol{s}_n}$ is a vector of coefficients describing the decomposition of the signal $\xv_n$ in terms of dictionary atoms.
Given the measurement set ${\bf Y}$ and the forward matrices $\{{\Psimat}_n\}_{n=1}^N$, we aim to jointly learn $\Dmat, \{{\boldsymbol{s}}_n\}_{n=1}^N$, and the noise precision parameter $\alpha_0$ to recover ${\bf X}$. One key difference of our model compared to other blind CS work~\cite{Gleichman11BCS} is that each patch has a unique $\boldsymbol{\Psi}_n$, which is inspired from our cameras since the mask is generated randomly and  $\boldsymbol{\Psi}_n$ also takes account of the system calibration.

We model each dictionary atom as a draw from a Gaussian distribution,
\begin{eqnarray}
\Dmat &=& \left[\dv_1, \dots, \dv_K \right], \nonumber \\
\dv_k &\sim& {\cal N}(0, \frac{1}{P} {\bf I}_P), ~~\forall k = 1,\dots, K.
\end{eqnarray}
The coefficients
$s_{k,n}$ are assumed drawn from the marginalized distribution
\begin{align}  \label{eq:double}
& p(s_{k,n}|\tau_n,\Phi_{k,n}) \nonumber\\
&=\int\hspace{-1mm}  {\cal N}(s_{k,n};0,\tau_n^{-1}\alpha^{-1}){\rm InvGa}(\alpha; 1,(2\Phi_{k,n})^{-1}) d \alpha\\
&= \frac{1}{2}\sqrt{\frac{\tau_n}{\Phi_{k,n}}}  \exp\left(-|s_{k,n}|\sqrt{\frac{\tau_n}{\Phi_{k,n}}}\right), \hspace{-1mm}\nonumber
\end{align}
where ${\rm InvGa}(\cdot)$ denotes the inverse-gamma distribution.
The parameter $\tau_n >0$ is a ``global'' scaling for all coefficients of the $n^{th}$ patch, and $\Phi_{k,n}$ is a ``local'' weight for the $k$th coefficient of that patch. 
We place a gamma prior ${\rm Ga}(g_0,h_0)$ on $\Phi_{k,n}$, where one may set the hyperparameters $(g_0,h_0)$ to ensure that most $\Phi_{k,n}$ are small. 
This encourages most $s_{k,n}$ to be small. 
By maximizing the log posterior to obtain a point estimate for the model parameters, one observes that the log of the prior in (\ref{eq:double}) corresponds to adaptive Lasso regularization \cite{ZouAdaLasso}.

Equivalently to (\ref{eq:double}), the model for $s_{k,n}$ may be represented in the hierarchical form
\begin{eqnarray}
s_{k,n}| \tau_{n},\alpha_{k,n}&\sim&\mathcal{N}(0,\tau_n^{-1}\alpha_{k,n}^{-1}), \label{Eq:Skn}\\
 \alpha_{k,n}|\Phi_{k,n} &\sim& \mbox{InvGa}(1,(2\Phi_{k,n})^{-1}),\label{eq:basic}\\
\tau_{n}&\sim& \mbox{Ga}(a_0,b_0),\\
\Phi_{k,n}&\sim& \mbox{Ga}(g_0,h_0), \label{eq:PhiGa}
\end{eqnarray}
where latent variables $\{\alpha_{k,n}\}$ are included in the generative model, instead of marginalizing them out as in (\ref{eq:double}). A vague/diffuse gamma prior is placed on the scaling parameters $\tau_n$. Despite introducing the latent variables $\{\alpha_{k,n}\}$, the form in (\ref{eq:basic}) is convenient for computation, and with an appropriate choice of $(g_0,h_0)$, most $\{\alpha_{k,n}\}$ are encouraged to be large. The large $\alpha_{k,n}$ corresponds to small $s_{k,n}$; this model imposes that most $s_{k,n}$ are small, $i.e.$ it imposes {\em compressibility}.
Note that (\ref{Eq:Skn}) is different from the model used in~\cite{Xing12SIAM}, where a single $\tau$ is used for all patches; here, we impose different compressibility for each patch by inferring a unique $\tau_n$, thus providing flexibility.

In order to automatically infer the number of necessary dictionary atoms, we can replace (\ref{eq:ywDs}) with
\begin{eqnarray}
\yv_n &=& {\boldsymbol{\Psi}}_n\Dmat {\boldsymbol \Lambda} \boldsymbol{s}_n + \epsilonv_n, \\
{\boldsymbol \Lambda} &=& {\rm diag}[\nu_1, \dots, \nu_K],\quad
\nu_k\sim{\cal N}(0, \eta_k^{-1}),\\
\eta_k &=& \prod_{j=1}^k \tilde{\eta}_j,  ~~\forall k=1,\dots,K;\quad
{\tilde\eta}_j \sim {\rm Ga}(e_0, f_0).
\end{eqnarray}
The multiplicative gamma prior (MGP)~\cite{Dunson11MultiGam} used above is developed to stochastically increase the precision $\eta_k$ as $k$ increases. 
During inference, we observe that as $k$ increases, $\nu_k$ tends to zero.
This results in an approximate ordering of $\nu_k$ by weight. 
Based on this weight, we can infer the importance of the columns of $\Dmat$.  
The other use of the $\nu_k$'s is to avoid over-fitting during the learning procedure. 
We can update a fraction of $\Dmat$ by the weight of $\nu_k$ and discard the atoms with small weights to reduce the computational cost (refer to the Appendix for the inferred $\nu_k$ and learned dictionary $\Dmat$).
\subsection{The Statistical Model}
The full statistical model of the proposed blind CS approach is:
\begin{eqnarray}
\yv_n &\sim& {\cal N}({\boldsymbol{\Psi}}_n\Dmat {\boldsymbol \Lambda} \boldsymbol{s}_n, \alpha_0^{-1}{\bf I}_P),\\
{\Dmat}&=& [\boldsymbol{d}_1,\dots, \boldsymbol{d}_K], ~~~{\boldsymbol{d}}_k \sim  {\cal N}(0, \frac{1}{P}{\bf I}_P),\\
s_{k,n} &\sim & {\cal N}(0, \tau_n^{-1} \alpha_{k,n}^{-1} \alpha_0^{-1}), \label{Eq:skn_alpha0}\\
\alpha_{k,n} &\sim& {\rm InvGa}(1, (2\Phi_{k,n})^{-1}),
\\
{\boldsymbol \Lambda} &=& {\rm diag}[\nu_1, \dots, \nu_K],\\
\nu_k &\sim&{\cal N}(0, \eta_k^{-1}), ~~
\eta_k = \prod_{j=1}^k \tilde{\eta}_j,\\
\tau_n &\sim& {\rm Ga}(a_0, b_0),~~~
\alpha_0 \sim {\rm Ga}(c_0, d_0),\\
{\tilde\eta}_j &\sim& {\rm Ga}(e_0, f_0),~~~{\Phi}_{k,n} \sim  {\rm Ga}(g_0, h_0),
\end{eqnarray}
where broad priors are placed on the hyperparameters $(a_0, b_0, c_0, d_0, e_0,f_0,g_0,h_0)$, {\em i.e.}, $a_0=\dots =h_0 = 10^{-6}$.
Note in (\ref{Eq:skn_alpha0}), the coefficients are also scaled to the noise precision $\alpha_0$.

Let ${\bf \Theta} = \{{\Dmat}, {\bf S}, {\bf \Lambda},{\boldsymbol \epsilonv}\}$ denote the parameters to be inferred.  The log of the joint posterior may be expressed as:
\begin{align}
&-\log p({\bf \Theta}|{\bf Y})  \nonumber \\
&= \frac{\alpha_0 \sum_{n}\|\yv_n - {\boldsymbol{\Psi}}_n\Dmat {\boldsymbol \Lambda} \boldsymbol{s}_n\|_2^2}{2} + \frac{P\sum_k\|\dv_k\|_2^2}{2}
+ \log {\rm Ga}(\alpha_0)
\label{eq:d}\\
&~ + \frac{\alpha_0 \sum_{n,k} \tau_n \alpha_{k,n} s_{k,n}^2}{2} + \sum_n \log {\rm Ga}(\tau_n) \nonumber\\
&~+\sum_{k,n}\log {\rm Ga}({\Phi}_{k,n}) + \sum_{n,k} \log {\rm InvGa}(\alpha_{k,n}; 1, \frac{1}{2\Phi_{k,n}}) \label{eq:s} \\
&~+ \frac{\sum_k \eta_k \nu_k^2}{2} + \sum_k \log {\rm Ga}({\tilde \eta}_k).  \label{eq:lambda}
\end{align}
The terms in (\ref{eq:d}) are widely employed in optimized-based dictionary learning~\cite{Aharon06TSP,Mairal07TIP,Elad06TIP}. 
The first term imposes an $\ell_2$ fit between the model and the observed data ${\yv_n}$ and the second term imposes an $\ell_2$ regularization on the dictionary atoms. 
The third term regularizes the relative importance of the aforementioned two terms in (\ref{eq:d}) via the weighting $\alpha_0$ (updated during the inference).
The terms in (\ref{eq:s}) impose both compressibility (via the global-shrinkage parameter $\tau_n$ and local-shrinkage parameter $\alpha_{k,n}$) and an $\ell_2$ regularization on $s_{k,n}$.
Finally, the terms in (\ref{eq:lambda}) impose both $\ell_2$ regularization and shrinkage on $\nu_k$.

\subsection{Related Models}
The shrinkage manifested by (\ref{eq:s}) is the most distinct aspect of the proposed model relative to
previous optimization-based (and Bayesian-based \cite{Zhou12TIP,Rajwade13SIAM}) approaches. 
These other approaches effectively impose sparsity through $\ell_1$ regularization on $s_n$. 
This is done by replacing terms in (\ref{eq:s}) with $\gamma_s \sum_n\|{\boldsymbol{s}}_n\|_1$.
To solve an optimization-based analog to the proposed approach, one seeks to minimize the objective function
\begin{eqnarray} \label{eq:opt}
{\cal L}(\bf \Theta) &=& \frac{\alpha_0 \sum_{n}\|\yv_n - {\boldsymbol{\Psi}}_n\Dmat {\boldsymbol \Lambda} \boldsymbol{s}_n\|_2^2}{2}\nonumber\\
 &&+ \frac{\sum_k\|\dv_k\|_2^2}{2} + \gamma_s \sum_n\|{\boldsymbol{s}}_n\|_1,
\end{eqnarray}
the parameters
$\alpha_0$ and  $\gamma_s$ are typically set by hand (e.g., via cross-validation).
One advantage of the Bayesian framework is that we infer posterior distributions for $\alpha_0$ and $\gamma_s$ (in our model this is ($\tau_n, \alpha_{k,n}$), along with similar posterior estimates for all model parameters without cross-validation.
We also note that if we replace (\ref{eq:s}) by the spike-slab prior as used in \cite{Zhou12TIP}, the model will reduce to BPFA. 
As opposed to this spike-slab prior, which imposes sparsity directly, our shrinkage prior imposes compressibility, and it is more appropriate to the high dimensional hyperspectral image patches. The {\em global-local} shrinkage prior used here can extract more useful information from the limited dictionary atoms\footnote{We did experiments of denoising and inpainting with benchmark color images and compared with K-SVD~\cite{Mairal07TIP} and BPFA~\cite{Zhou12TIP}. The results are shown in the appendix. The proposed algorithm constantly performs better than the above two methods.}.

Other forms of shrinkage priors like Gaussian scale model and Laplace scale mode can be found in~\cite{Armagan12,Carvalho_horseshoe,garrigues2010group,griffin2011bayesian}. Aiming to better mimic the marginal behavior of discrete mixture priors, the global-local shrinkage priors~\cite{Polson10shrinkglobally,Polso12Levy} have been developed to offer sufficient flexibility in high-dimensional settings, which inspires our model.

\subsection{Inference}
Due to local conjugacy, we can write the conditional posterior distribution for all parameters of our model in closed form, making the following Markov Chain Monte Carlo (MCMC) inference based on Gibbs sampling a straightforward procedure. 
\begin{enumerate}
\item[1)]
Sampling ${\boldsymbol{d}}_k$:
\begin{eqnarray} 
p({\boldsymbol{d}}_k|-) &\propto& {\cal N}({\boldsymbol{\mu}}_{d_k},{\bf \Sigma}_{d_k}); \\
{\bf \Sigma}_{d_k} &=& \left[P {\bf I}_P + \alpha_0 \nu_k^2 \sum_{n=1}^N s_{k,n}^2 {\boldsymbol{\Psi}}_n^T{\boldsymbol{\Psi}}_n\right]^{-1}, \nonumber\\
{\boldsymbol{\mu}}_{d_k} &=& \alpha_0 {\bf \Sigma}_{d_k} \nu_k\sum_{n=1}^N s_{k,n} {\boldsymbol{\Psi}}_n^T {\boldsymbol{y}}_{n,-k}, \nonumber\\
{\boldsymbol{y}}_{n,-k} &=& {\boldsymbol{y}_n - {\boldsymbol{\Psi}}_n \Dmat\boldsymbol{\Lambda s}}_n +  {\boldsymbol{\Psi}}_n{\boldsymbol{d}}_k \nu_k s_{k,n}. \nonumber
\end{eqnarray}
\item[2)]
Sampling $s_{k,n}$:
\begin{eqnarray}
p(s_{k,n}|-) &\propto & {\cal N}(\mu_{s_{k,n}},\sigma^2_{s_{k,n}}); \\
\sigma^2_{s_{k,n}}&=& \left(\tau_{n}\alpha_{k,n}\alpha_{0} + \alpha_{0} \lambda_k^2{\boldsymbol{d}}_k^{T}{\boldsymbol{\Psi}}_n^T{\boldsymbol{\Psi}}_n{\boldsymbol{d}}_k\right)^{-1}, \nonumber\\
\mu_{s_{k,n}}&=& \alpha_{0}\sigma^2_{s_{k,n}} \nu_k{\boldsymbol{d}_k}^T {\boldsymbol{\Psi}}_n^T{\boldsymbol{y}}_{n,-k}. \nonumber
\end{eqnarray} 
\item[3)]
Sampling $\tau_n$:
\begin{equation}
p(\tau_n|-) \propto {\rm Ga}\left(a_0 + \frac{1}{2}K , b_0 + \frac{1}{2} \sum_{k=1}^K s_{k,n}^2 \alpha_{k,n}\alpha_{0}\right).
\end{equation}
\item[4)]
Sampling $\alpha_{k,n}$:
\begin{eqnarray}
p(\alpha_{k,n}|-)
&\propto & {\rm IG}\left(\sqrt{\frac{1}{\Phi_{k,n}s_{k,n}^2\tau_{n}\alpha_{0}}}, \frac{1}{\Phi_{k,n}}\right),
\end{eqnarray}
where ${\rm IG}(\cdot)$ denotes the inverse-Gaussian distribution.
\item[5)]
Sampling $\Phi_{k,n}$:
\begin{eqnarray}
p(\Phi_{k,n}|-) &\propto& {\rm GIG}(2h_0, \alpha_{k,n}^{-1}, g_0 -1),
\end{eqnarray}
where ${\rm GIG}(x; a,b,p)$ is the generalized inverse Gaussian distribution
\begin{equation}
{\rm GIG}(x;a,b,p) = \frac{(a/b)^{\frac{p}{2}}}{2 K_p(\sqrt{ab})} x^{p-1}\exp\left(-\frac{1}{2}(ax + \frac{b}{x})\right), \nonumber
\end{equation}
and $K_p(\theta)$ is the modified Bessel function of the second kind
\begin{equation}
K_p(\theta) = \int_0^{\infty} \frac{1}{2}\theta^{-p} t^{p-1}\exp\left(-\frac{1}{2}(t+\frac{\theta^2}{t})\right) dt.\nonumber
\end{equation}

\item[6)]
Sampling $\alpha_{0}$:
\begin{eqnarray}
p(\alpha_{0}|-) 
&\propto& {\rm Ga}(c_1, d_1); \\
c_1 &=& c_0 + \frac{1}{2}\sum_n \|{\boldsymbol{\Psi}}_n\|_0 + \frac{1}{2}KN, \nonumber\\
d_1 &=& d_0 + \frac{1}{2}\sum_{n}\|{\boldsymbol{y}}_n - {\boldsymbol \Psi}_n \Dmat {\boldsymbol{ \Lambda s}}_n\|_2^2 \nonumber \nonumber \\
&&+ \frac{1}{2}\sum_n \sum_k s_{k,n}^2 \tau_n \alpha_{k,n}, \nonumber
\end{eqnarray}
where $\|{\boldsymbol{\Psi}}_n\|_0$ denotes the number of nonzero entries in ${\boldsymbol{\Psi}}_n$, and ``-" refers to the conditioning
parameters of the distributions.
\end{enumerate}
The update equations of MGP related parameters $\{\nu_k,\eta_k,{\tilde\eta}_k\}$ can be found in~\cite{Dunson11MultiGam} (also listed in the Appendix).
In applications where speed is important, we can use all conditional posteriors including those above
to derive a variational Bayes (VB) inference algorithm for our model, which loosely amounts to
replace conditioning variables with their corresponding moments. Details of the VB procedure can be found in the Appendix.

\begin{table*}[htbp!]
\caption{Reconstruction PSNR (dB) of various scenes with different algorithms. Each feature shows the mean value and standard derivation of the PSNRs of reconstructed hyperspectral images across wavelength channels.}
\centering
\begin{tabular}{|l|c|c|c|c|c|}
\hline Scene 
&  \includegraphics[height = 2cm, width=2cm]{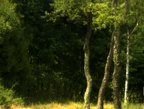}  &\includegraphics[height = 2cm, width=2cm]{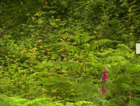}& \includegraphics[height = 2cm, width=2cm]{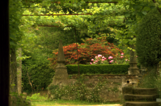} & \includegraphics[height = 2cm, width=2cm]{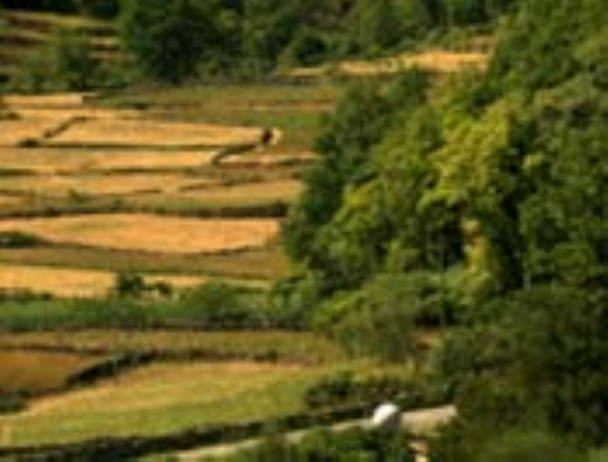} & \includegraphics[height = 2cm, width=2cm]{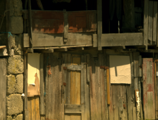}  \\
\hline \hline
 Shrinkage w/o RGB & 30.9673$\pm$5.0328 & 26.6319$\pm$5.5290 & 32.8013$\pm$4.1525 & 42.9004$\pm$5.4524  & 34.5532$\pm$5.1652 \\
\hline Shrinkage w/ RGB & {\bf 32.6175$\pm$4.0417}  & {\bf 34.0775$\pm$4.3469} & {\bf 35.1950$\pm$2.9645} & {\bf 47.5275$\pm$4.7856}   & {\bf 35.8182$\pm$2.0722}\\
\hline TwIST & 21.8238$\pm$2.9894  & 23.2787$\pm$5.8818 & 25.2030$\pm$3.4791  & 19.9874$\pm$3.7625    & 21.6566$\pm$2.2755\\
\hline Bregman & 30.7027$\pm$4.6553  & 27.2907$\pm$5.4581 & 32.8595$\pm$3.8964  & 27.1390$\pm$4.0307  & 34.0448$\pm$4.5355 \\
\hline
\end{tabular}
\label{Table:HSI_PSNR}
\end{table*}
\subsection{RGB Images as Side Information}
While reconstructing the hyperspectral images is challenging (the compression ratio is $N_{\lambda}:1$ when a single measurement is available), an RGB image of the same scene can be obtained easily by an off-the-shelf color camera in the unused path of the system (i.e. directly reflecting off the polarizing beamsplitter) as shown in Fig. \ref{fig:System}. We here consider using the RGB image as side information to aid the reconstruction.

In the case of an additional side RGB camera, the measurement is a joint dataset composed of the RGB image and the CASSI measurement; the compression ratio can be considered as $(N_{\lambda} +3):4$. The new measurement model is:
\begin{eqnarray} \label{eq:side}
\underbrace{\left[\begin{array}{l}
{\bf Y} \\
{\bf Y}^{(\rm rgb)}
\end{array}\right]}_{\stackrel{\rm def}{=} {\tilde {\bf Y}}} &=& \left[\begin{array}{cc}
\Psimat & {\bf 0} \\
{\bf 0} & {\bf I}
\end{array} \right] \underbrace{\left[ \begin{array}{l}
\Xmat \\
\Xmat^{(\rm rgb)}
\end{array}\right]}_{\stackrel{\rm def}{=} {\tilde \Xmat}} + \epsilonv,
\end{eqnarray}
where $\{{\bf Y}^{(\rm rgb)}, \Xmat^{(\rm rgb)}\}$ are the patch format RGB image. 
Considering each patch,
\begin{eqnarray}
\left[\begin{array}{l}
{\yv}_n \\
{\yv}^{(\rm rgb)}_n \end{array}\right]&=& \left[\begin{array}{cc}
\Psimat_n & {\bf 0} \\
{\bf 0}& {\bf I} \end{array}\right] \left[\begin{array}{l}
\xv_n\\
\xv_n^{(\rm rgb)}
\end{array} \right] + {\tilde \epsilonv}_n,
\end{eqnarray}
where $\{{\yv}_n^{(\rm rgb)}\}_{n=1}^N $ is the vectorized form of the RGB image patch corresponding to $\{{\yv}_n\}_{n=1}^N$ and similar for $\{{\xv}_n^{(\rm rgb)}\}_{n=1}^N $.
The dictionary learning method proposed in Section~\ref{Sec:DL} is now performed on the joint dataset ${\tilde \Xmat} = \tilde{\bf D}  \tilde{\bf S}$ to learn the super-dictionary $\tilde{\Dmat} \in {\mathbb R}^{{\tilde P}\times K}$, where $\tilde{P} = P + 3n_xn_y$.

Given ${\tilde {\bf Y}}$, the proposed model learns $\tilde{\bf D}$ and  $\tilde{\bf S}$ simultaneously.
For the recovery of hyperspectral images, we select the top $P$ rows from the product of $\tilde{\bf D}$ and  $\tilde{\bf S}$. Averaging the overlapping patches and reformatting the data to 3D give us the desired hyperspectral images.
We aim to get both clear images at each wavelength and correct spectra for each pixel, which are both embedded in the dictionary atoms.
Since the RGB images are fully observed and embed the same scene as the hyperspectral images, this side information will facilitate the algorithm to learn a better dictionary and therefore improving the reconstruction quality.

Another way to use the RGB image is treating each R, G and B channel as a superposition of the hyperspectral images with different weights corresponding to the quantum efficiency of the R, G, and B sensors. 
However, these quantum efficiencies may be different for each camera.
Here, we aim to propose a general and robust framework to collaborate RGB images with CASSI measurements. Therefore, the formulation in (\ref{eq:side}) is adopted.

In our experimental setup, a grayscale image may also be used as side information.
As the RGB camera is inexpensive and carrying richer spectral information, we only consider the RGB case in the following experiments.
In the results presented below, we consider that RGB image is measured separately by an additional camera. We are now modifying our SLM-CASSI system to capture the RGB image and the compressed hyperspectral image simultaneously as illustrated in Fig.~\ref{fig:System}.

\section{Experimental Results \label{Sec:Results}}
In this section, we verify the proposed blind CS inversion algorithm on diverse datasets.  
For now, we solely consider the case where  a single measurement ({\em i.e.}, the compressed hyperspectral image) is available\footnote{Please note that~\cite{Rajwade13SIAM} did not show the algorithm proposed therein working with a single measurement. In this paper, we focus on the single measurement case and our model can readily be used in multiple measurements scenario. We also tried the method proposed in~\cite{Rajwade13SIAM} and the results are worse than ours.}. 
The proposed blind CS algorithm is compared with:
$i$) two-step iterative shrinkage/thresholding (TwIST) \cite{Bioucas-Dias2007TwIST} (using a 2D total variation norm on images at each wavelength\footnote{We also tried the 3D total variation (TV) on the entire datacube, the results are a little bit worse than the 2D TV. Therefore, we only show the 2D TV results of TwIST.}),
and $ii$) the linearized Bregman algorithm \cite{Yin08bregman}.
The linearized Bregman approach respectively regularizes the spectral dimension and spatial dimensions with $\ell_1$-norms of the discrete cosine transformation and wavelet transformation coefficients.
Note that TwIST and linearized Bregman are {\em global} algorithms; it is necessary to load the entire data into RAM when performing inversion.  Conversely, our proposed method is {\em locally}-(patch) based; the online learning is feasible and the inversion can be performed in batch-mode~\cite{MairalICML09}.

We employ a spatial patch size $n_x = n_y = 8$.  The compression ratio, $N_{\lambda}$, depends on the dataset. The proposed model can learn the dictionary atom number $K$ from the data. During experimentation, we have found that setting $K=64$ usually provides good results.  We have verified that further increasing $K$ does not significantly change the outcome of our methods. 
All code used in the experiments was written in Matlab and executed on a 3.3GHz desktop with 16GB RAM.

We evaluate the proposed algorithm on synthetic data presented in Section~\ref{Sec:SimData} and the real data captured both by the CASSI~\cite{Wagadarikar09CASSI,Kittle10AO} and the proposed SLM-CASSI camera Section~\ref{Sec:RealData}.  We denote our method as ``shrinkage" in the experiments (in figures and ta bles) as the shrinkage prior is used in our model.  To evaluate the algorithm's performance, we use the PSNR of the reconstructed images at different wavelengths and the correlation between the reconstructed spectrum and the true (reference) spectrum.

The computational time of our model is similar to the BPFA model used in~\cite{Rajwade13SIAM} and linearized Bregman. TwIST provides faster, but generally worse results than our algorithm in terms of the metrics mentioned above.  Quantitatively, linearized Bregman and TwIST require about 20 minutes and 14 minutes, respectively, to reconstruct the bird data (size $768\times 1024\times 24$).

\begin{figure}[ht!]
      \centering
       \includegraphics[scale = 0.6]{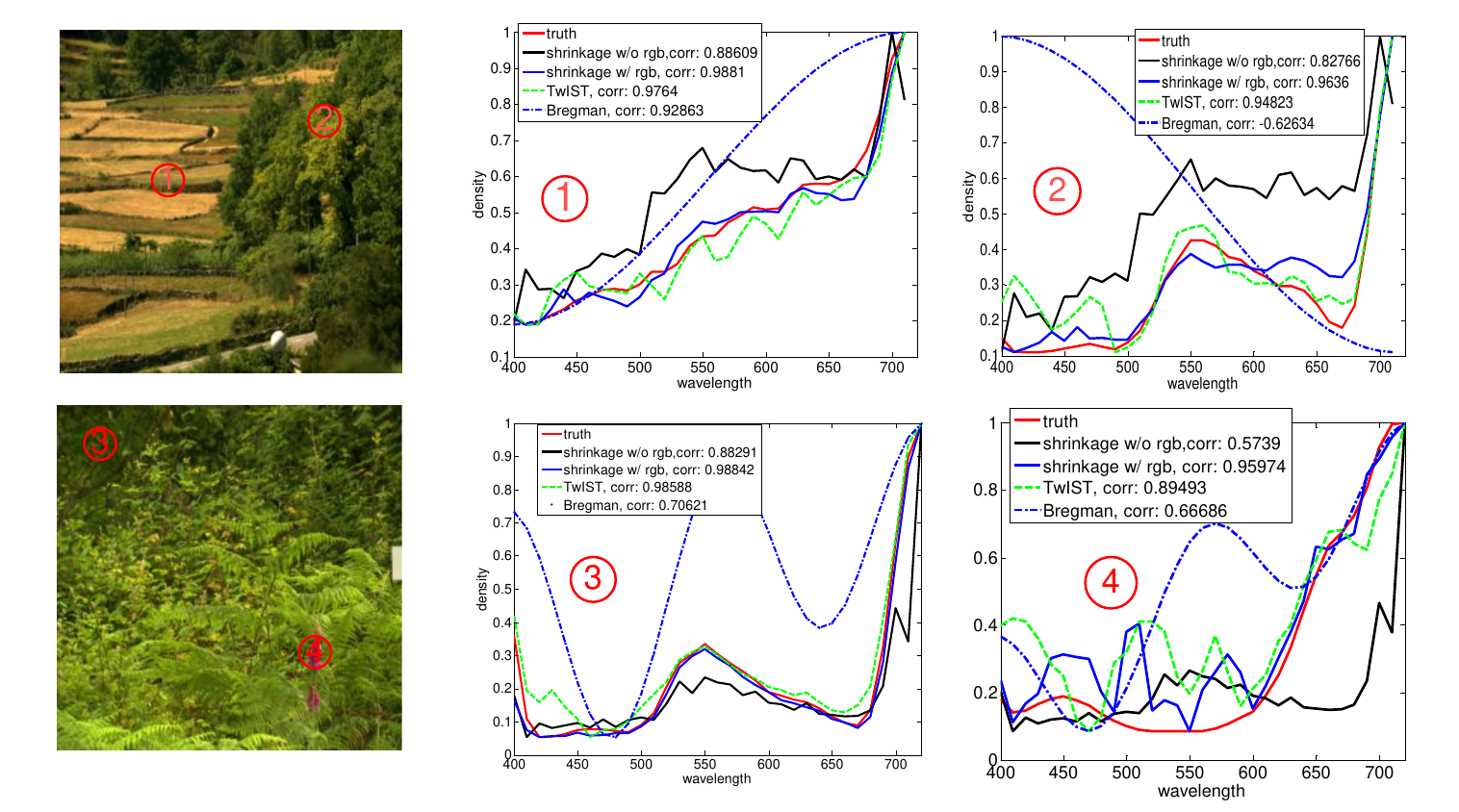}\\
       \caption{Reconstructed spectra with different algorithms compared to ground truth. ``corr" denotes the correlation between the reconstructed spectra and the truth.}
      \label{Fig:spectrum_sim_scene}
      \end{figure}
\subsection{Simulation Data \label{Sec:SimData}}
We use hyperspectral images encoded with a random binary (Bernoulli(0.5)) mask  to simulate the CASSI measurements.  The RGB images are available and aligned well with the hyperspectral images.

\subsubsection{Nature scenes}
We first consider the hyperspectral images of natural scenes used in~\cite{Foster06JOSA}\footnote{\url{http://personalpages.manchester.ac.uk/staff/d.h.foster/Hyperspectral_images_of_natural_scenes_04.html}}. There are $N_{\lambda} = 33$ channels (400-720nm with a 10nm interval) and we resize each image to $N_x = N_y =512$.  The PSNR of each reconstructed spectral channel with mean values and standard deviations (across wavelength channels) are presented for each algorithm in Table~\ref{Table:HSI_PSNR}.
It can be seen that the proposed shrinkage method outperforms the others, especially once the RGB images are used as side information.
Fig.~\ref{Fig:spectrum_sim_scene} plots the reconstructed spectrum of selected blocks (the average spectrum of pixels inside the block is used) in the scene compared to the ground truth.
It can be seen that:
$i$) though TwIST provides the lowest PSNR, the reconstructed spectrum is usually correct;
$ii$) the spectrum reconstructed by the proposed algorithm is improved significantly when side information is provided;
$iii$) the linearized Bregman presents the worst spectrum, mainly due to the DCT used in the spectral domain for inversion.
\begin{figure}[ht!]
       \centering
       \includegraphics[scale = 0.8]{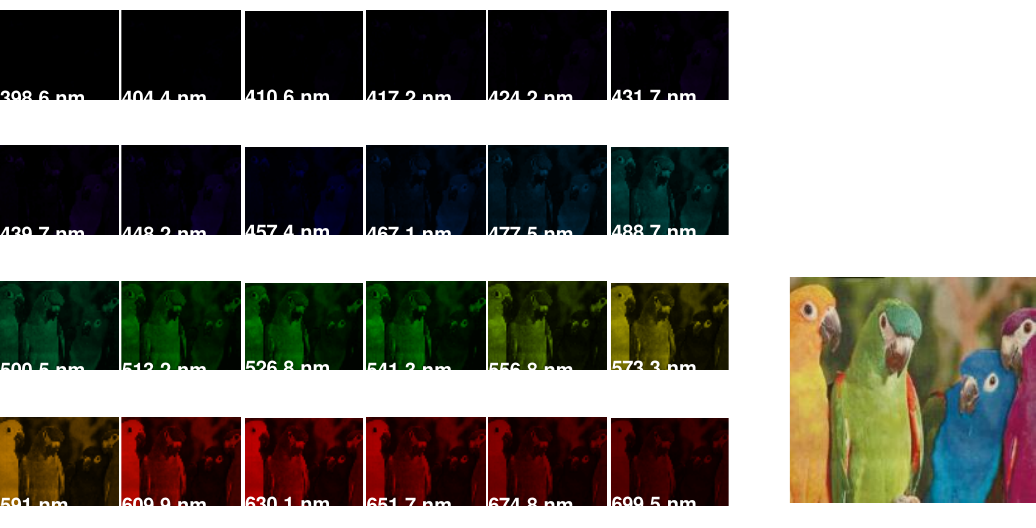}\\
       \caption{Left: the hyperspectral images (ground truth), right: the RGB image.}
       \label{Fig:Bird_truth}
   \end{figure} 
\subsubsection{Bird data}
\begin{figure}[ht!]
       \centering
       \includegraphics[scale = 0.6]{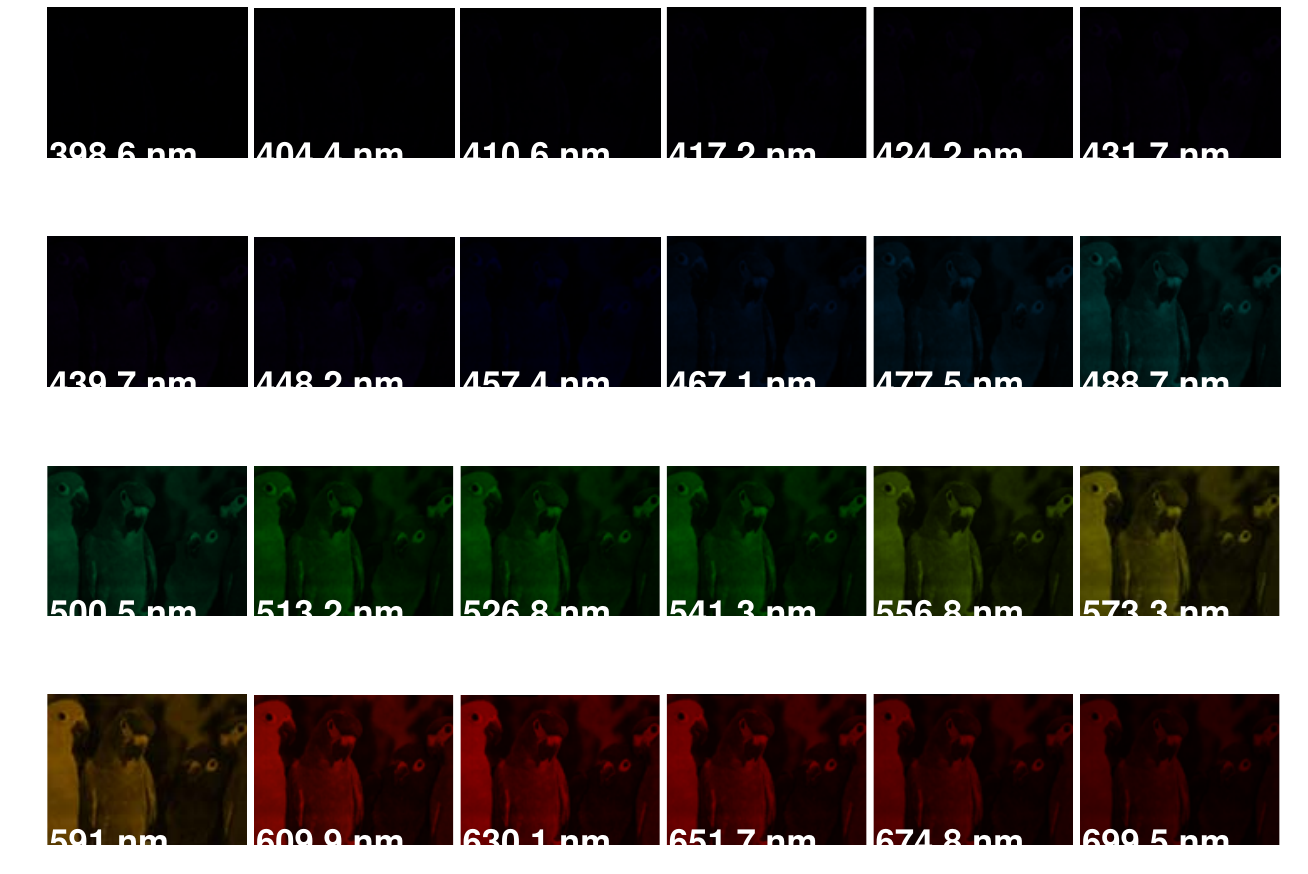}\\
       \vspace{0.5cm}
       \includegraphics[scale = 0.6]{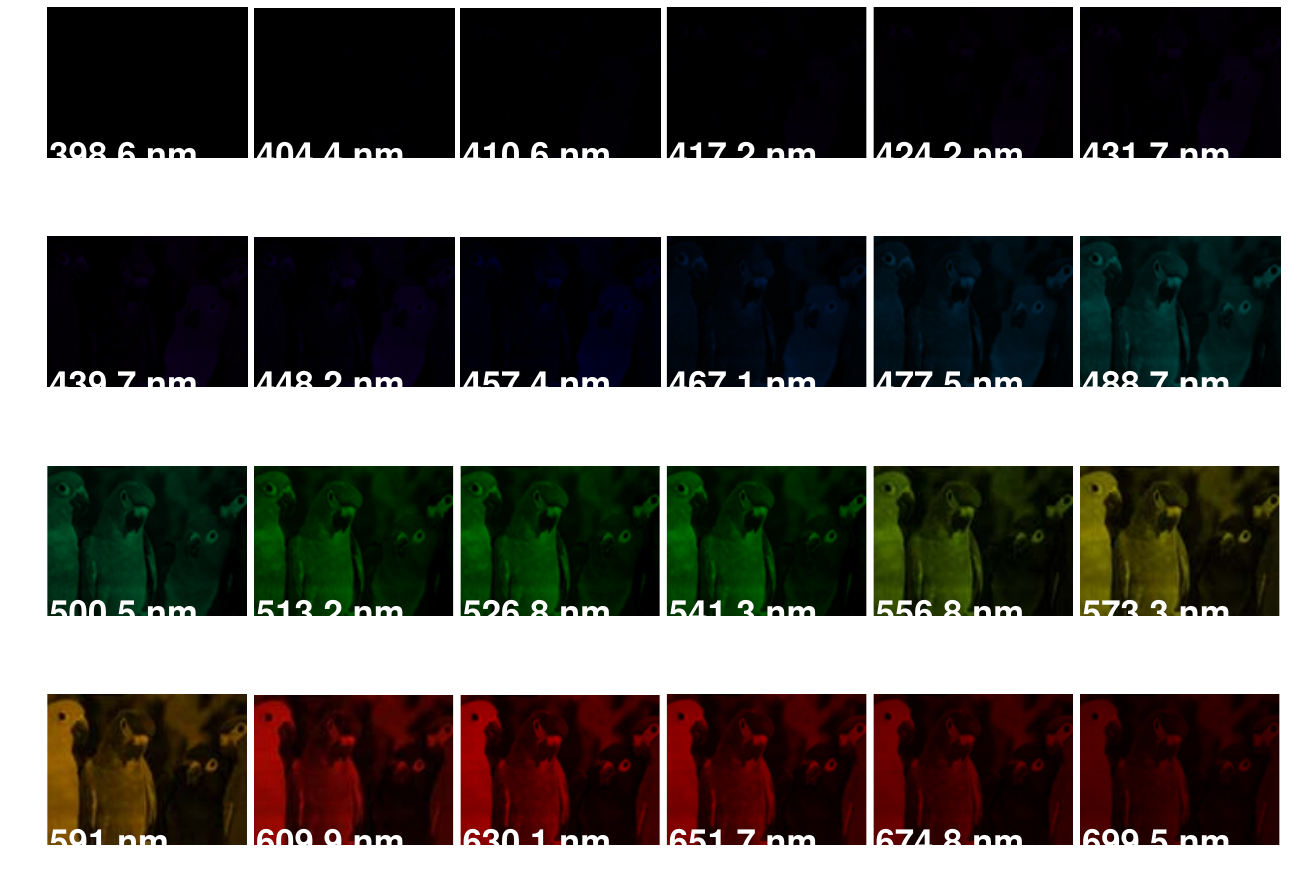}
       \caption{Reconstructed images without RGB image (top) and with RGB image (bottom).}
       \label{Fig:Bird_rec_sim}
   \end{figure}

\begin{figure}[ht!]
          \centering
      \includegraphics[scale = 0.22]{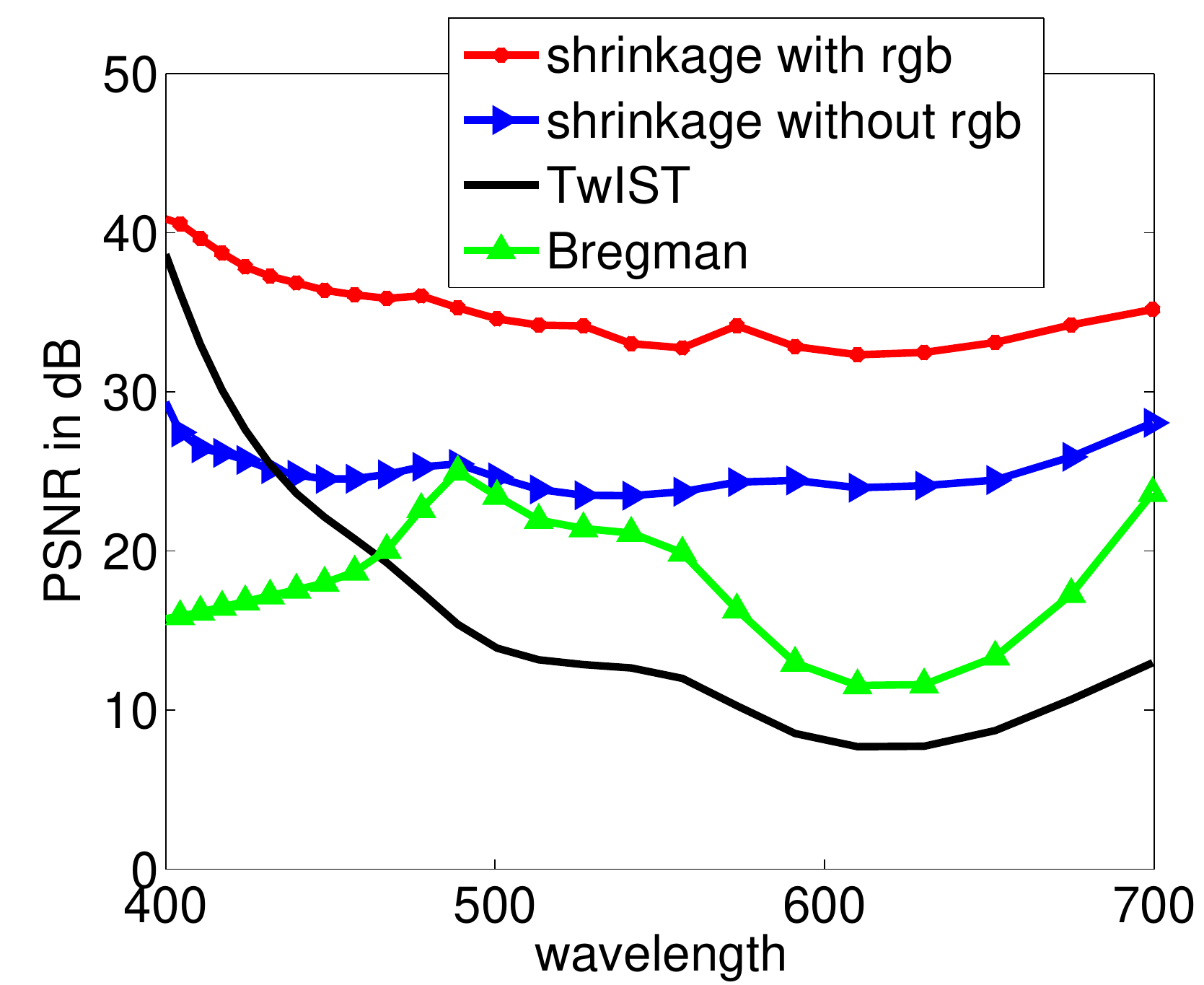}\includegraphics[scale =0.25]{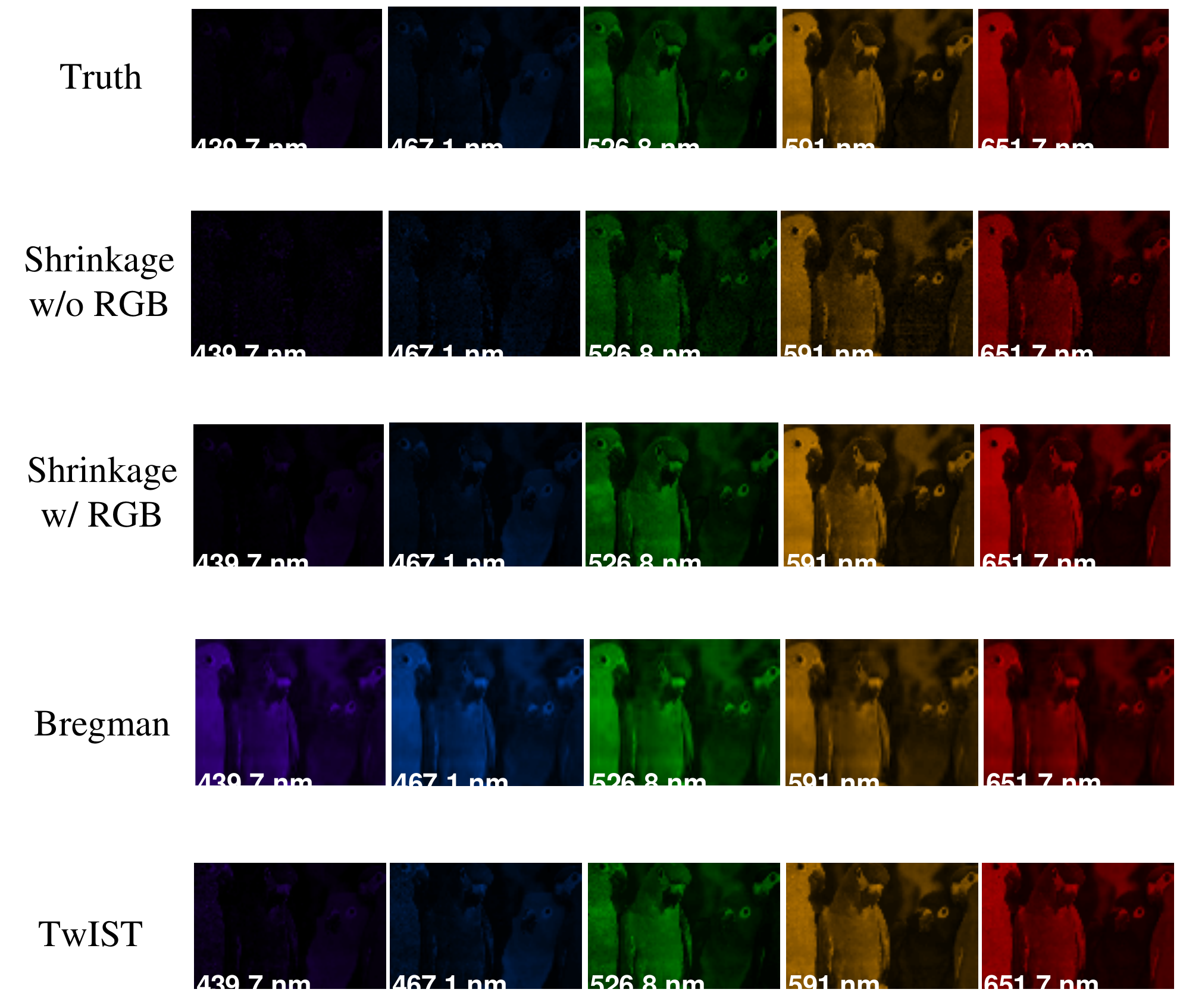}
          \caption{Left: PSNRs of the reconstructed images at each wavelength using different methods. Right: selected channels of reconstructed images compared to truth. }
          \label{Fig:Bird_zoom_sim}
      \end{figure}

\begin{figure}[ht!]
       \centering
       \includegraphics[scale = 0.4]{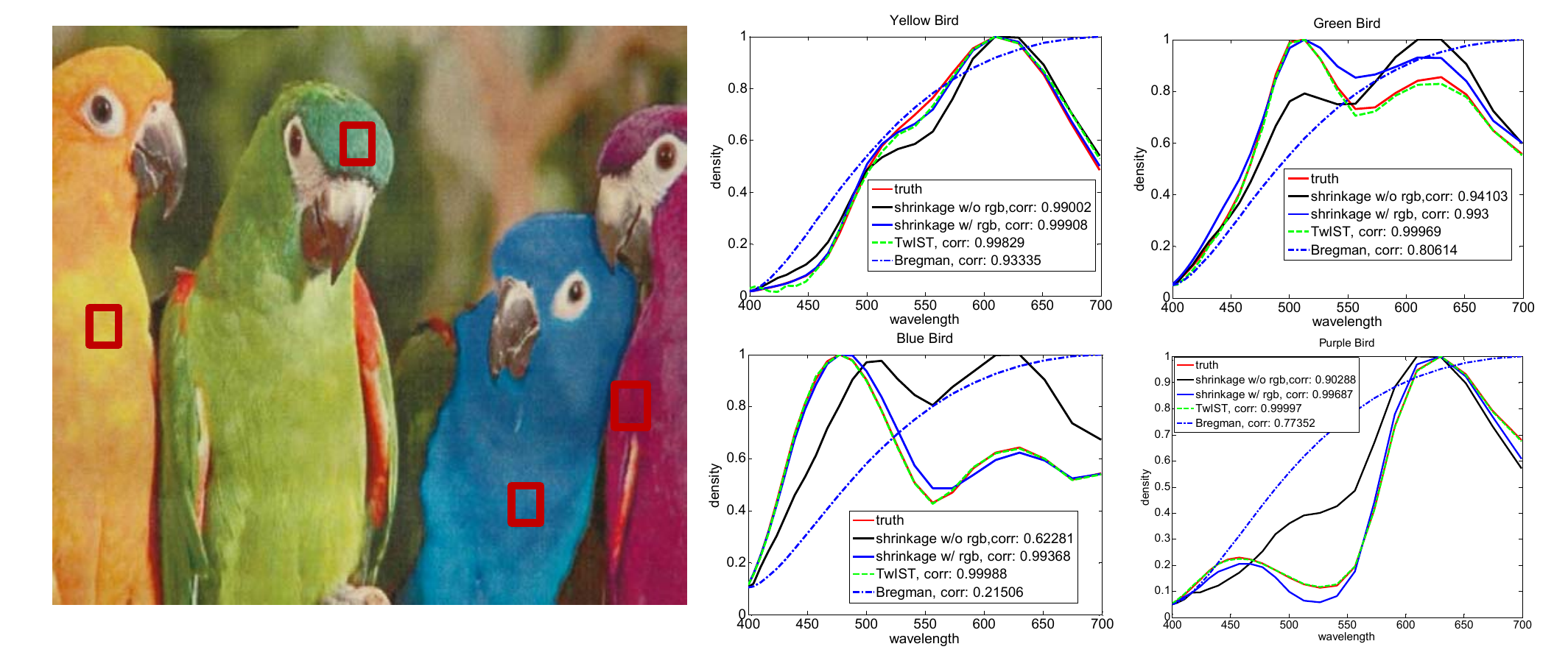}
       \caption{Reconstructed spectra of different birds with various algorithms.}
       \label{Fig:Spec_comp_bird_sim}
   \end{figure}

Next we consider the 24-channel bird data (Fig.~\ref{Fig:Bird_truth}) measured by a hyperspectral camera in~\cite{Rajwade13SIAM}.
The RGB image is also available.
Fig.~\ref{Fig:Bird_rec_sim} shows the reconstructed images using the proposed algorithm with and without side information.
The left part of Fig.~\ref{Fig:Bird_zoom_sim} plots the reconstruction PSNR of each channel using different algorithms. Again, our shrinkage blind CS method coupled with the RGB image provides the best result.
The right part of Fig.~\ref{Fig:Bird_zoom_sim} shows the reconstructed images of five selected channels using the different algorithms. It can be seen that the TwIST results are characterized by lost details (over smoothed) of the birds; our model's results without using the RGB images appear noisy.
Fig.~\ref{Fig:Spec_comp_bird_sim} depicts the reconstructed spectra of different birds with different algorithms.
It can be seen that both TwIST and our algorithm with the RGB image provide very good matches to the ground truth, while the proposed model without RGB images and linearized Bregman do not represent the spectrum well.

\begin{figure*}[ht!]
       \centering
       \includegraphics[scale = 0.3]{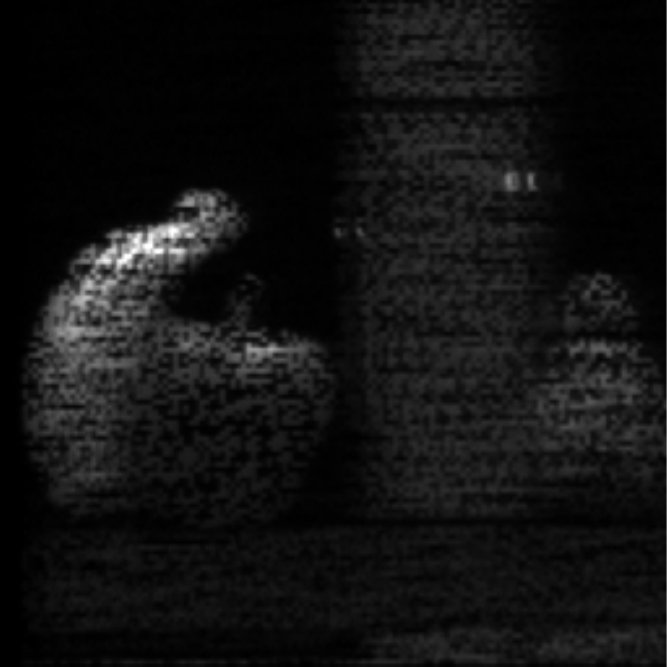} \hspace{1cm}
       \includegraphics[scale = 0.36]{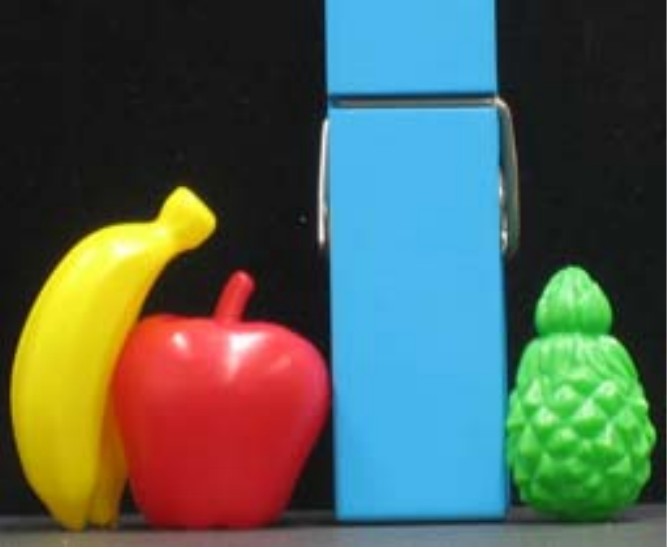}\\
       \includegraphics[scale = 1.0]{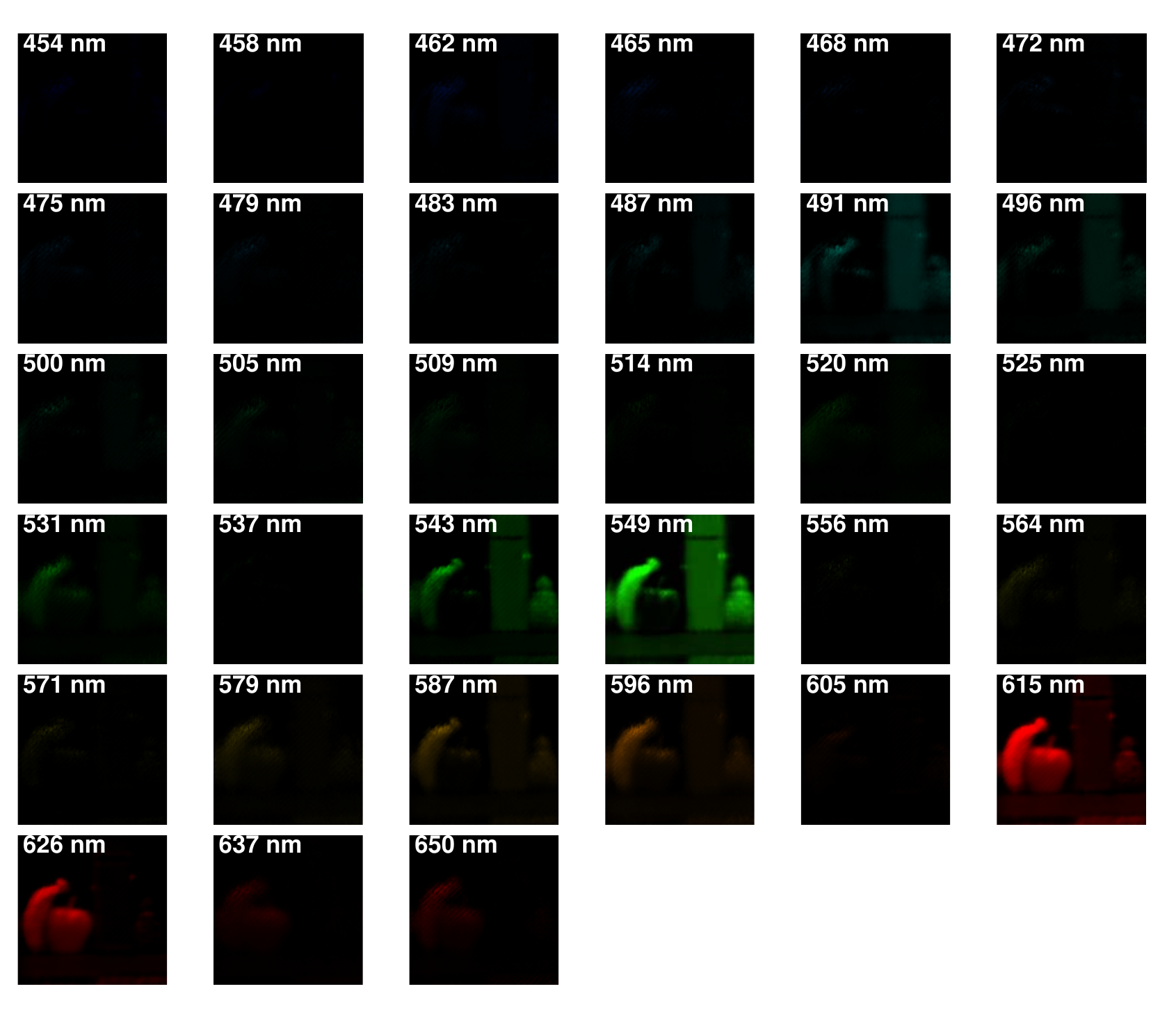}
       \caption{Real data: measurement and reconstruction without side information (the RGB image).}
       \label{Fig:Object_mea_recon}
   \end{figure*} 
 \begin{figure}[ht!]
           \centering
           \includegraphics[scale =0.6]{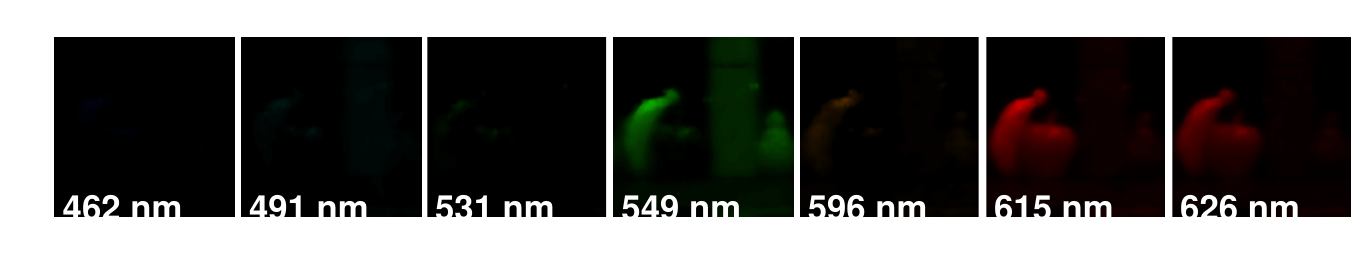}\\
           \hspace{1mm}\includegraphics[scale =0.64]{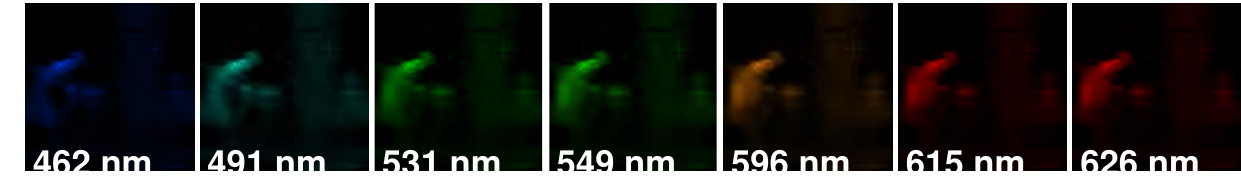}\\
           \includegraphics[scale =0.6]{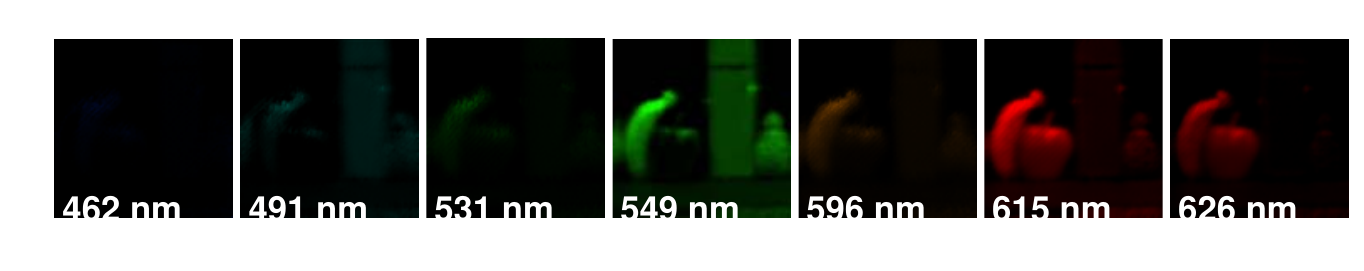}
           \caption{Real data: selected channels of reconstructed hyperspectral images. Top row: TwIST, middle row: linearized Bregman, bottom row: shrinkage without RGB. Notice that only the bottom row provides the clear ``apple stem".}
           \label{Fig:object_zoom}
       \end{figure} 
 \begin{figure}[ht!]
           \centering
       \includegraphics[scale = 0.2]{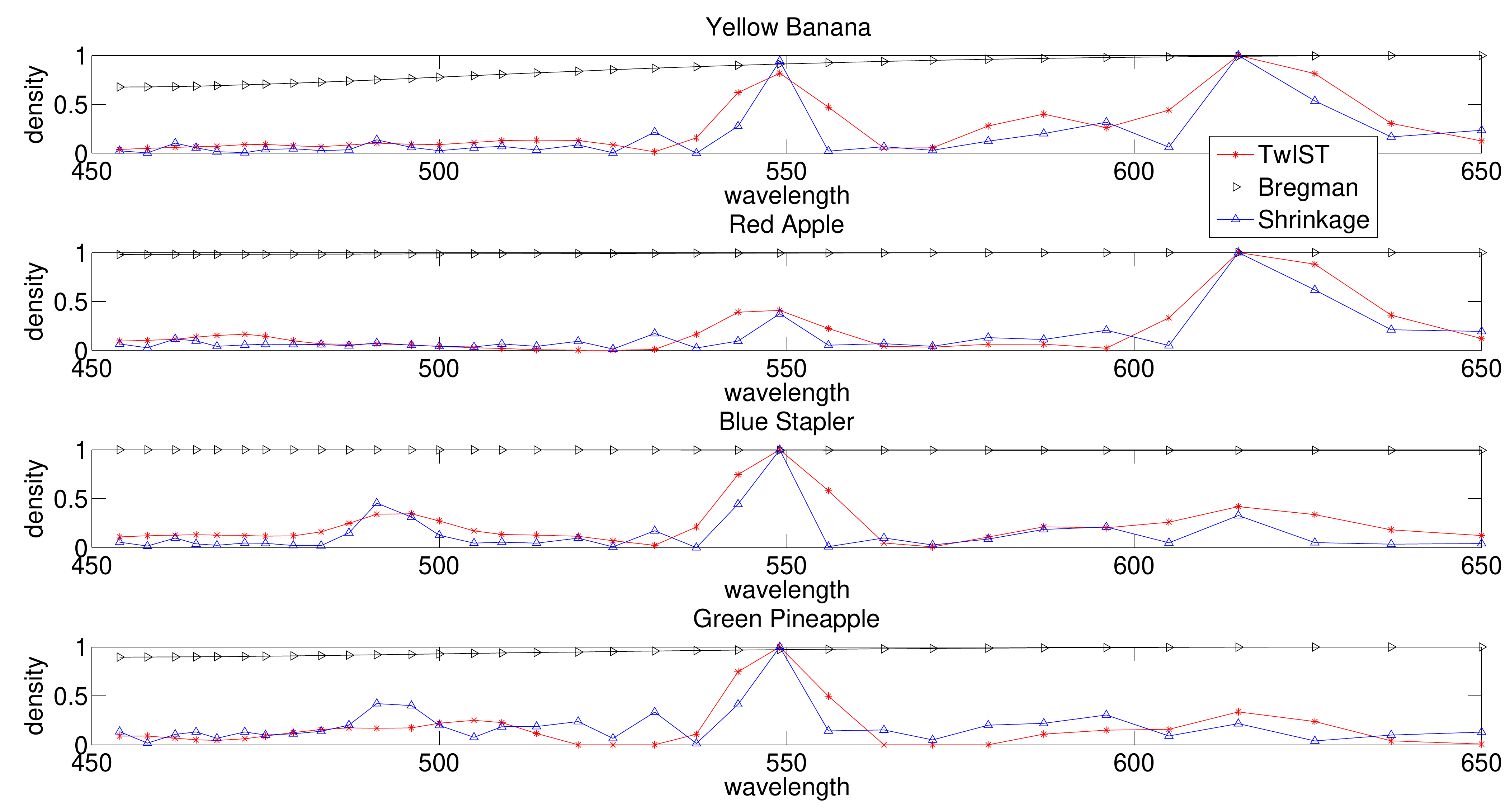}
           \caption{Real data: spectra of the reconstructed hyperspectral images for the object data.}
           \label{Fig:object_spec}
       \end{figure}        
       
\subsection{Real Data \label{Sec:RealData}}
In this section, we apply the proposed algorithm to real data captured by our cameras (both the original CASSI camera and the new SLM-CASSI camera).
\subsubsection{Object data}
We first demonstrate our algorithm on data taken by the original CASSI system~\cite{Wagadarikar09CASSI}\footnote{\url{http://www.disp.duke.edu/projects/CASSI/experimentaldata/index.ptml}}.  In these experiments, the reconstructions have 33 spectral channels (marked on the reconstruction in Fig.~\ref{Fig:Object_mea_recon}) and $N_x = N_y = 256$.
There are 4 objects in the scene, a red apple, a yellow banana, a blue stapler and a green pineapple.
Fig.~\ref{Fig:Object_mea_recon} shows the reconstruction of the proposed algorithm without the RGB image. Since the RGB image is not well-aligned with the CASSI measurement, we don't show the result with side information and we found the reconstruction with the RGB image (not aligned) looks similar to this one due to the simple scene used in this experiment.
Fig.~\ref{Fig:object_zoom} compares selected reconstructed images inverted by different algorithms. It can be seen that the proposed algorithm provides more detail of the scene (notice the clear apple stem).
Fig.~\ref{Fig:object_spec} plots the reconstructed spectra of the four objects. TwIST~\cite{Wagadarikar09CASSI} has yielded accurate spectra; our algorithm (without side information) provides similar results. The linearized Bregman algorithm does not reconstruct this real data well; we don't not show its results in the following experiments.

\subsubsection{Bird data}
\begin{figure}[ht!]
       \centering
       \includegraphics[scale = 0.15]{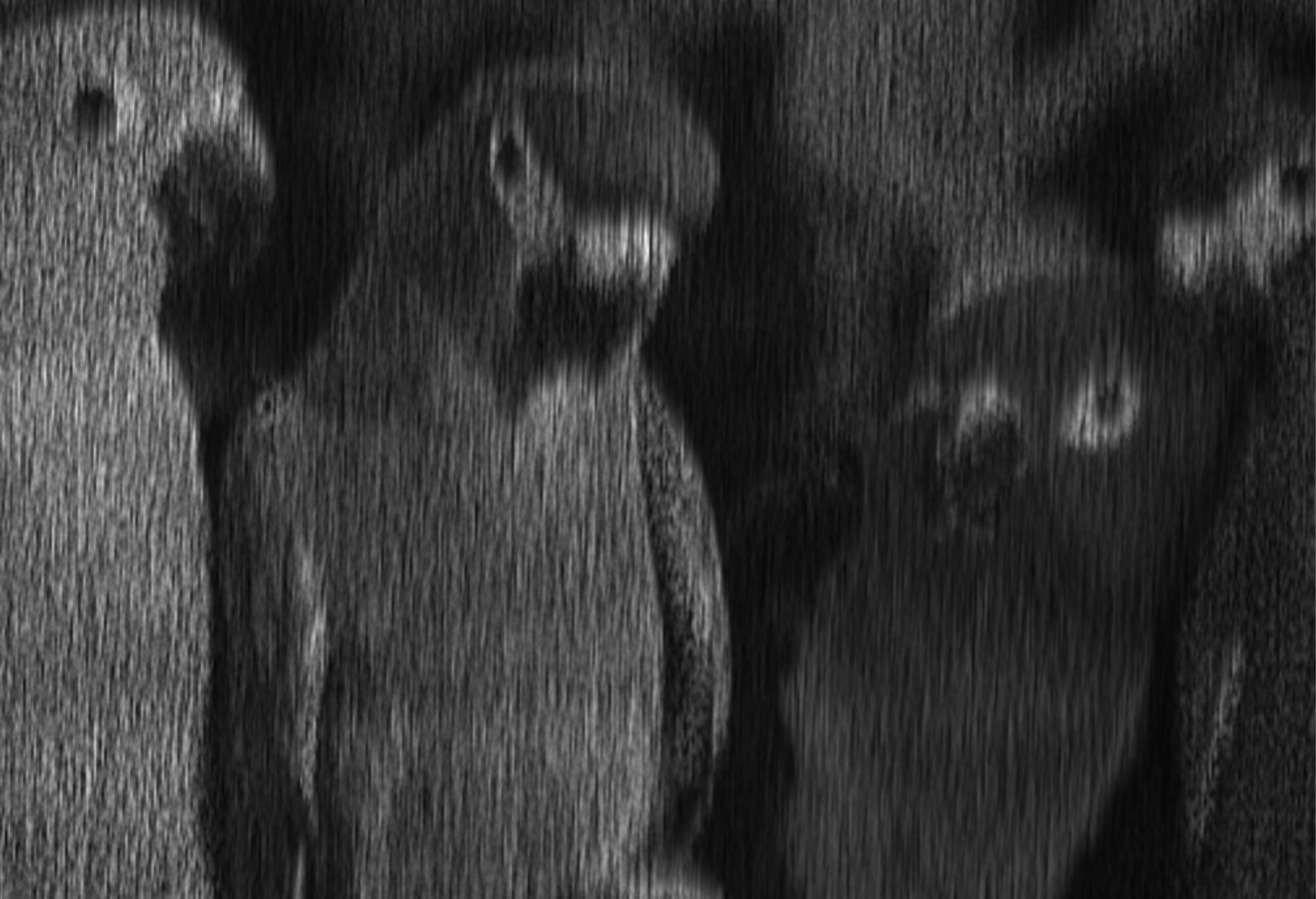}\\
       \includegraphics[scale = 0.4]{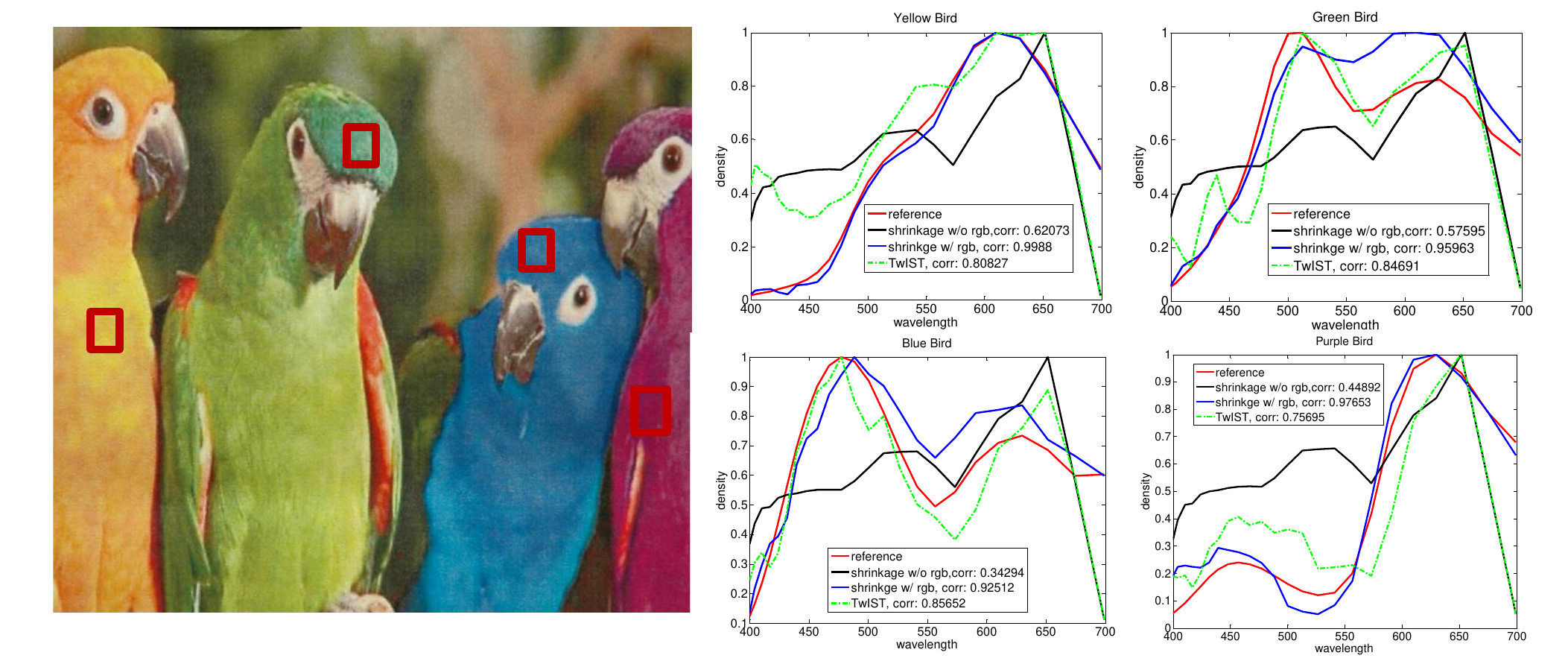}
       \caption{Real data: Top: measurement; bottom: reconstructed spectra of different birds with various algorithms compared to references.}
       \label{Fig:Spectrum_real_bird}
   \end{figure}

\begin{figure*}[ht!]
       \centering
       \includegraphics[scale = 1.3]{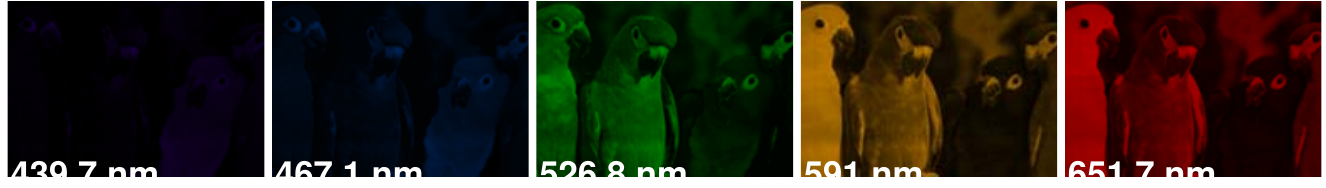}\\
       \includegraphics[scale = 1.3]{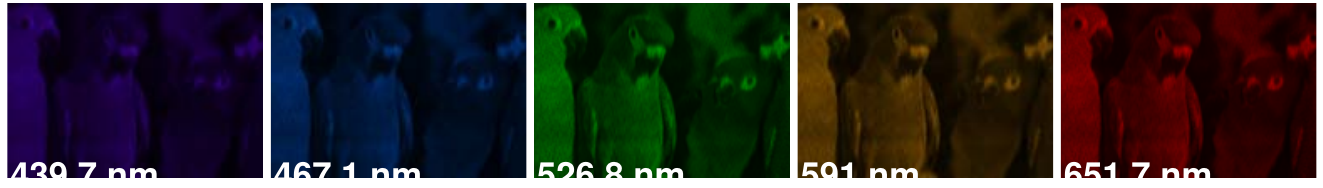}\\
       \includegraphics[scale = 1.3]{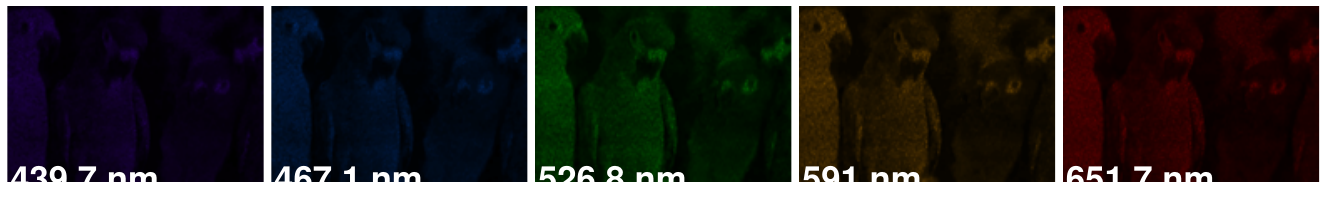}\\
       \includegraphics[scale = 1.3]{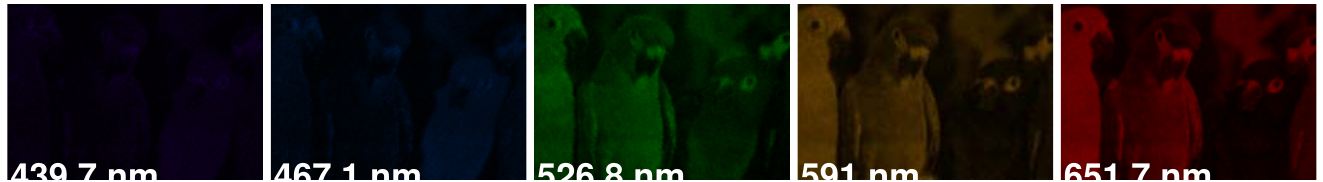}
       \caption{Real data, row 1: reference; and reconstruction with, row 2: shrinkage without the RGB image; row 3: TwIST; row 4: shrinkage with the RGB image. Notice the blue bird in the first two images for a good  spectrum recovery.}
       \label{Fig:img_real_bird}
   \end{figure*}

We again consider the bird data, now captured by the multiframe-CASSI camera~\cite{Kittle10AO}. The RGB image is aligned manually with the CASSI measurement ($N_x =703, N_y = 1021, N_{\lambda} = 24$).
We plot the spectra in Fig.~\ref{Fig:Spectrum_real_bird}; the reconstructed images are shown in Fig.~\ref{Fig:img_real_bird}.
It can be seen clearly that our proposed method with side information provides the best results with respect to both image clarity and spectral accuracy.  Without the use of side information, the proposed model yields clear images but reconstructs the spectra poorly. 
This verifies the benefit of the side information in real data.

\subsubsection{Data with CASSI-SLM }
\begin{figure}[ht!]
       \centering
       \includegraphics[scale = 0.6]{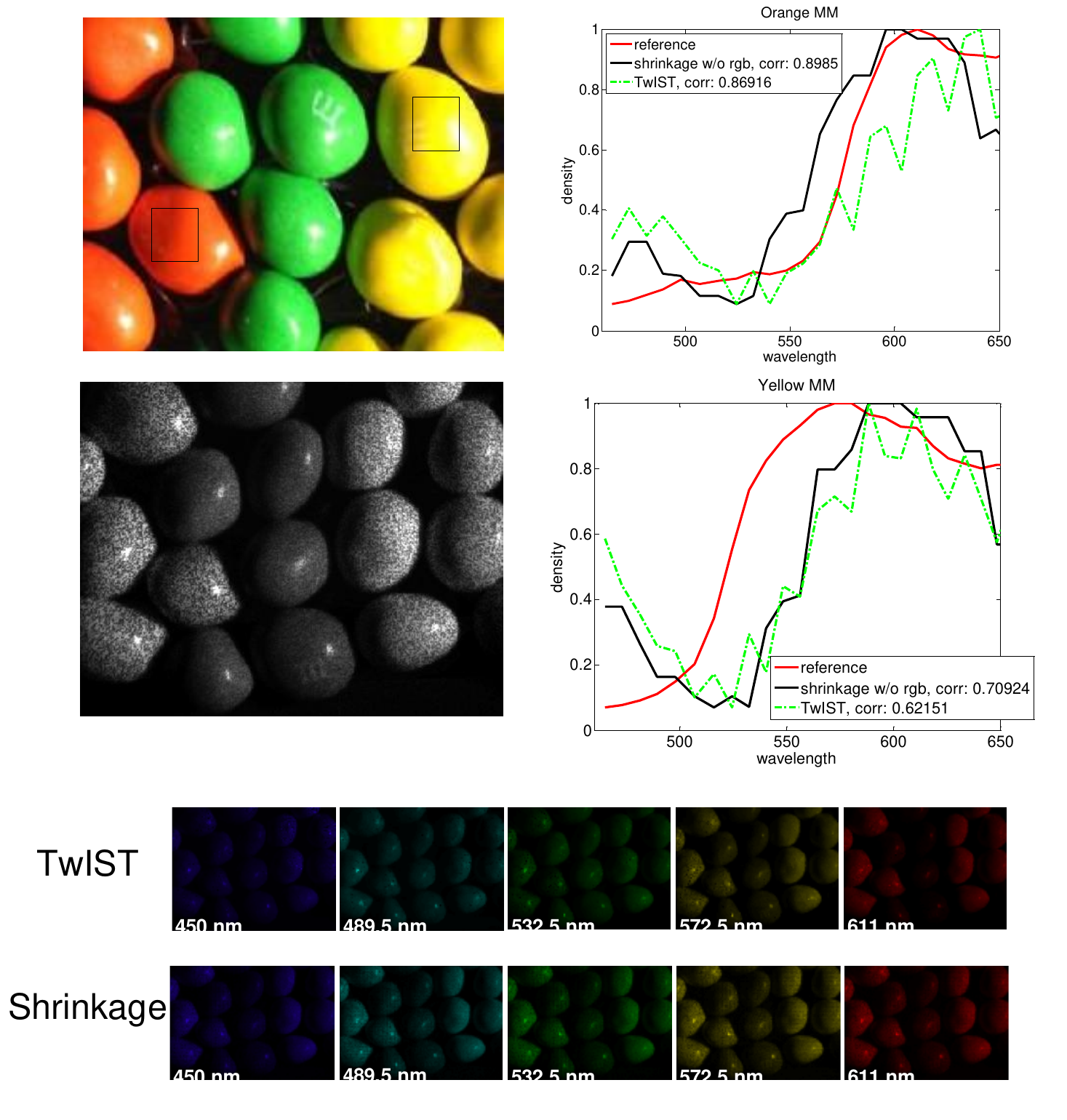}
       \caption{Real data: reconstructed spectra and images of M\&M with TwIST and shrinkage without RGB image.}
       \label{Fig:MM}
   \end{figure}

Now we consider the dataset captured by the proposed camera.
Since no RGB image is available, we only show the results of our algorithm without side information. Fig.~\ref{Fig:MM} shows the reconstructed spectrum and selected frames for the M\&M dataset ($N_x =512, N_y = 784, N_{\lambda} = 30$)\footnote{The 30 wavelengths are 450nm, 458nm, 465nm, 473nm, 481.5nm, 489.5nm, 498nm, 507nm, 516nm, 524.5nm, 532.5nm, 540.5nm, 548.5nm, 556,5nm, 564.5nm, 572.5nm, 580.5nm, 588.5nm, 596nm, 603.5nm, 611nm, 618.5nm, 625.8nm, 633.5nm, 641nm, 648.5nm, 656nm, 663.5nm, 671nm, 678.5nm.}.  As with the original CASSI experiments, we use a fiber optic spectrometer  (USB2000, Ocean Optics) to provide the reference spectra for the targets.
It can be seen again our algorithm provides better results than TwIST, both for images and spectrum.

\begin{figure*}[ht!]
       \centering
       \includegraphics[scale = 0.9]{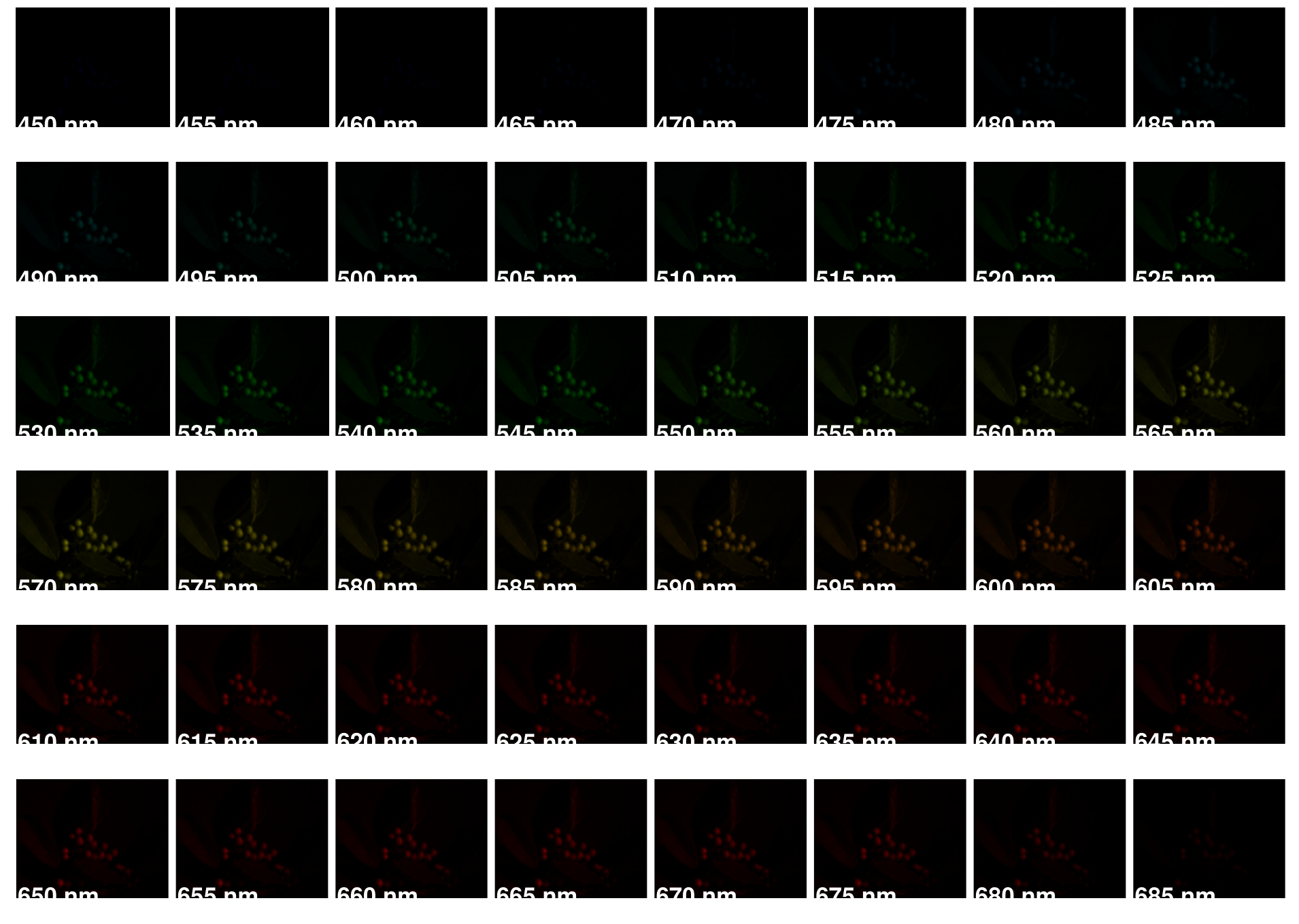}
       \caption{Real data: reconstructed images of Berry data without RGB image.}
       \label{Fig:Berry_real}
   \end{figure*}

As a last example, we show the reconstructed images with our method for the berry data ($N_x = N_y = 2048, N_{\lambda} = 40$) in Figure~\ref{Fig:Berry_real}.
Notice that the leaf reconstructs are prevalent in the green channels (520$\sim$590nm), while the berries (red) appear in the red portion of the spectrum (610$\sim$680nm).

\section{Conclusions \label{Sec:Col}}
We have developed and tested a new Bayesian dictionary learning model for blind CS.  Specifically, we have demonstrated high-quality inversion of compressed hyperspectral images captured by real cameras.
Via the global-local shrinkage priors, our algorithm imposes compressibility on the dictionary coefficients to extract more information from the dictionary atoms.
The reconstruction quality is improved significantly by integrating the compressed measurements with RGB images as side information under the blind CS framework.
We also developed a new compressive hyperspectral imaging camera that uses an SLM to perform the spectral coding. Experimental results demonstrate the feasibility of the new camera, the superior performance of the algorithm, and the benefit of side information.

The original CASSI camera utilizes a separate disperser and encoder; the SLM-CASSI is capable of modulating both dimensions with one element. 
The SLM-CASSI also provides the potential of video rate compressive sensing and multi-frames adaptive sensing using the 60 Hz refresh rate of the SLM. 
SLM-CASSI offers greater flexibility of masking functions and an additional RGB camera for side information; however, it intertwines spectral and spatial multiplexing capabilities.  The coded aperture/disperser architecture used in the original CASSI camera is more reliable, has a smaller form factor (i.e. fewer optical parts), and is less expensive, but lacks the coding flexibility and ability to use a separate camera for side information.

In this paper, we have shown that by using the RGB image as side information in our proposed blind CS framework, the reconstruction quality of hyperspectral images can be improved significantly.  This framework can be generalized to other applications and algorithms.  For instance, Gaussian mixture model based dictionary learning approaches~\cite{Chen10SPT,Yang14GMM,Yu11SPT,Yang14GMM2,Yu12IPT} that learn a union of subspaces can also benefit from side information.

\bibliographystyle{IEEEtran}

\clearpage
\newpage
\section{Model and Inference}
\label{Sec:Model}
The full statistical model is:
\begin{eqnarray}
\yv_n &\sim& {\cal N}({\boldsymbol{\Psi}}_n\Dmat {\boldsymbol \Lambda} \boldsymbol{s}_n, \alpha_0^{-1}{\bf I}_P),\\
{\Dmat}&=& [\boldsymbol{d}_1,\dots, \boldsymbol{d}_K], \\
{\boldsymbol{d}}_k &\sim & {\cal N}(0, \frac{1}{P}{\bf I}_P), \\
s_{k,n} &\sim & {\cal N}(0, \tau_n^{-1} \alpha_{k,n}^{-1} \alpha_0^{-1}),\label{eq:Skn}\\
\alpha_{k,n} &\sim& {\rm InvGa}(1, (2\Phi_{k,n})^{-1}),\\
{\Phi}_{k,n} &\sim & {\rm Ga}(g_0, h_0),\\
\tau_n &\sim& {\rm Ga}(a_0, b_0),\\
\alpha_0 &\sim &{\rm Ga}(c_0, d_0).\\
{\boldsymbol \Lambda} &=& {\rm diag}[\nu_1, \dots, \nu_K],\\
\nu_k&\sim&{\cal N}(0, \eta_k^{-1}), \\
\eta_k &=& \prod_{j=1}^k \tilde{\eta}_j, \\
{\tilde\eta}_j &\sim& {\rm Ga}(e_0, f_0).
\end{eqnarray}
The posterior density function of model
parameters may be represented as:
\begin{eqnarray}
p({\bf \Theta}|\boldsymbol{Y})&\propto& \prod_{n=1}^N{\cal N}\left[(\yv_n;{\boldsymbol{\Psi}}_n\Dmat {\boldsymbol \Lambda} \boldsymbol{s}_n, \alpha_0^{-1}{\bf I}_P) {\rm Ga}(\tau_n; a_0,b_0)  \right] \nonumber\\
&&\times \prod_{k=1}^K\left[{\cal N}(\dv_k; 0 ,\frac{1}{P}{\bf I}_P) {\cal N}(\nu_k; 0, \eta_k^{-1})\right]\nonumber\\
&&\times \prod_{k=1}^K \left[\prod_{n=1}^N{\cal N}(s_{k,n}; 0, \tau_n^{-1}\alpha_{k,n}^{-1}\alpha_0^{-1})\right] \nonumber\\
&&\times \prod_{k=1}^K\left[{\rm Ga}({{\tilde \eta}_k};e_0,f_0)  {\rm Ga}(\Phi_{k,n}; g_0, h_0)\right]\nonumber\\
&&\times \prod_{k=1}^K\left[ \prod_{n=1}^N{\rm InvGa}(\alpha_{k,n}; 1, (2\Phi_{k,n})^{-1}) \right]\nonumber\\
&&\times {\rm Ga}(\alpha_0; c_0, d_0).
\end{eqnarray}

\subsection{MCMC Inference}
\begin{enumerate}
\item[1)]
Sampling ${\boldsymbol{d}}_k$:
\begin{eqnarray} 
p({\boldsymbol{d}}_k|-) &\propto& {\cal N}({\boldsymbol{\mu}}_{d_k},{\bf \Sigma}_{d_k}); \\
{\bf \Sigma}_{d_k} &=& \left[P {\bf I}_P + \alpha_0 \nu_k^2 \sum_{n=1}^N s_{k,n}^2 {\boldsymbol{\Psi}}_n^T{\boldsymbol{\Psi}}_n\right]^{-1},\\
{\boldsymbol{\mu}}_{d_k} &=& \alpha_0 {\bf \Sigma}_{d_k} \nu_k\sum_{n=1}^N s_{k,n} {\boldsymbol{\Psi}}_n^T {\boldsymbol{y}}_{n,-k}, \\
{\boldsymbol{y}}_{n,-k} &=& {\boldsymbol{y}_n - {\boldsymbol{\Psi}}_n \Dmat\boldsymbol{\Lambda s}}_n +  {\boldsymbol{\Psi}}_n{\boldsymbol{d}}_k \nu_k s_{k,n}.
\end{eqnarray}
\item[2)]
Sampling $s_{k,n}$:
\begin{eqnarray}
p(s_{k,n}|-) &\propto & {\cal N}(\mu_{s_{k,n}},\sigma^2_{s_{k,n}}); \\
\sigma^2_{s_{k,n}}&=& \left(\tau_{n}\alpha_{k,n}\alpha_{0} + \alpha_{0} \nu_k^2{\boldsymbol{d}}_k^{T}{\boldsymbol{\Psi}}_n^T{\boldsymbol{\Psi}}_n{\boldsymbol{d}}_k\right)^{-1}, \\
\mu_{s_{k,n}}&=& \alpha_{0}\sigma^2_{s_{k,n}} \nu_k{\boldsymbol{d}_k}^T {\boldsymbol{\Psi}}_n^T{\boldsymbol{y}}_{n,-k}.
\end{eqnarray} 
\item[3)]
Sampling $\tau_n$:
\begin{equation}
p(\tau_n|-) \propto {\rm Ga}\left(a_0 + \frac{1}{2}K , b_0 + \frac{1}{2} \sum_{k=1}^K s_{k,n}^2 \alpha_{k,n}\alpha_{0}\right).
\end{equation}
\item[4)]
Sampling $\alpha_{k,n}$:
\begin{equation}
p(\alpha_{k,n}|-)
\propto  {\rm IG}\left(\sqrt{\frac{1}{\Phi_{k,n}s_{k,n}^2\tau_{n}\alpha_{0}}}, \frac{1}{\Phi_{k,n}}\right),
\end{equation}
where ${\rm IG}$ denotes inverse-Gaussian distribution.
\item[5)]
Sampling $\Phi_{k,n}$:
\begin{eqnarray}
p(\Phi_{k,n}|-) &\propto& {\rm GIG}(2h_0, \alpha_{k,n}^{-1}, g_0 -1),
\end{eqnarray}
where ${\rm GIG}(x: a,b,p)$ is the generalized inverse Gaussian distribution:
\begin{equation}
{\rm GIG}(x;a,b,p) = \frac{(a/b)^{\frac{p}{2}}}{2 K_p(\sqrt{ab})} x^{p-1}\exp\left(-\frac{1}{2}(ax + \frac{b}{x})\right), \nonumber
\end{equation}
and $K_p(\theta)$ is the modified Bessel function of the second kind
\begin{equation}
K_p(\theta) = \int_0^{\infty} \frac{1}{2}\theta^{-p} t^{p-1}\exp\left(-\frac{1}{2}(t+\frac{\theta^2}{t})\right) dt.\nonumber
\end{equation}
\item[6)]
Sampling $\alpha_{0}$:
\begin{eqnarray}
p(\alpha_{0}|-) 
&\propto& {\rm Ga}(c_1, d_1); \\
c_1 &=& c_0 + \frac{1}{2}\sum_n \|{\boldsymbol{\Psi}}_n\|_0 + \frac{1}{2}KN,\\
d_1 &=& d_0 + \frac{1}{2}\sum_{n}\|{\boldsymbol{y}}_n - {\boldsymbol \Psi}_n \Dmat {\boldsymbol{ \Lambda s}}_n\|_2^2 \nonumber\\
&&+ \frac{1}{2}\sum_n \sum_k s_{k,n}^2 \tau_n \alpha_{k,n},
\end{eqnarray}
where $\|{\boldsymbol{\Psi}}_n\|_0$ denotes the number of nonzero entries in ${\boldsymbol{\Psi}}_n$.
\item[7)]
Sampling $\nu_k$:
\begin{eqnarray}
p(\nu_k|-) & \propto & {\cal N}(\mu_{\nu_k}, \sigma^2_{\nu_k}); \\
 \sigma^2_{\nu_k} &=& \left(\eta_k + \alpha_0  \sum_{n=1}^N s_{k,n}^2 \dv_k^T {\boldsymbol{\Psi}}_n^T {\boldsymbol{\Psi}}_n\dv_k \right)^{-1},\\
\mu_{\nu_k} &=& \sigma^2_{\nu_k} \alpha_0 \dv_k^T \sum_{n=1}^N {\boldsymbol{\Psi}}_n^T \yv_{n,-k} s_{k,n}.
\end{eqnarray}
\item[8)]
Sampling $\tilde{\eta}_{j}$:
\vspace{-5mm}
\begin{eqnarray}
p(\tilde{\eta}_{j}) &\propto& {\rm Ga}(e_1,f_1);\\
e_1 &=& e_0+ \frac{1}{2}(K+1-j),\\
f_1 &=& f_0 + \frac{1}{2}\sum_{q=j}^{K}\nu_q^2\frac{\eta_q}{\tilde{\eta}_j}.
\end{eqnarray}
\end{enumerate}

\subsection{Variational Bayesian Inference} 
We adopt a mean-field variational Bayesian (VB) inference in lieu of its improved runtime.
A VB approach attempts to approximate the posterior distribution by a simpler distribution, $p({\bf \Theta}|{\boldsymbol Y}) \approx q({\bf \Theta})$,
where ${\bf Y}$ is the observed data matrix and ${\bf \Theta}$ denotes the set of independent latent variables in the model~\cite{Beal03VB}. VB assumes a complete factorization across latent variables, $q({\bf \Theta}) = \prod_i q_i({\bf \Theta}_i)$.
We define ${\bf \Theta} = \{{\boldsymbol D}, {\boldsymbol S}, {\boldsymbol \Lambda},{\boldsymbol \epsilonv}\}$ for the purpose of this work.

Solving for the optimal distribution $q^{\star}({\bf \Theta})$ that minimizes the distance between $p$ and $q$ effectively estimates the conditional posterior distribution $p({\bf \Theta}|{\bf Y})$.  A commonly-used distance metric between the two distributions functions is the Kullback-Leibler (KL) divergence~\cite{KL51}. We write the KL-divergence of $p$ from $q$ as follows:
\begin{equation}
{\rm KL}(q\|p) = \int_{{\bf \Theta}} q({\bf \Theta}) \ln \frac{q({\bf \Theta})}{p({\bf \Theta}|{\boldsymbol Y})}d{\bf \Theta},
\end{equation}
which can be simplified to
\begin{eqnarray}
\ln p({\bf Y})& =& {\rm KL}(q\|p) + \L(q), \\
\L(q)& =& -\int_{\bf \Theta} q({\bf \Theta})\ln\frac{q({\bf \Theta})}{p({\bf \Theta},{\boldsymbol Y})}d{\bf \Theta}.
\end{eqnarray}
We here observe that $\ln p({\bf Y})$ is fixed with respect to the variations in $q({\bf \Theta})$. Therefore, maximizing the Evidence Lower Bound (ELBO) $\L(q)$ is equivalent to minimizing the KL-divergence between the two distributions. This minimal distance occurs when
\begin{equation}
\ln q^{\star}({\bf \Theta}) = {\mathbb E}[\ln p({\boldsymbol Y}, {\bf \Theta})] + {\rm const}.
\end{equation}
Assuming a complete factorization across the latent variables $q({\bf \Theta}) = \prod_i q_i({\bf \Theta}_i)$, each parameter in a variational Bayes model is independently updated according to
\begin{equation} \label{eq:VB}
q_j^{\star}({\bf \Theta}_j) \propto \exp \{{\mathbb E}_{i\neq j}[\ln p({\boldsymbol Y}, {\bf \Theta})]\}.
\end{equation}

The update equations of VB are straightforward from the posterior distribution of Gibbs sampling ($\langle \cdot \rangle$ represents the expectation of the random variable inside):
\begin{align*}
\langle \dv_k\rangle &= \langle \alpha_0 \rangle \langle {\bf \Sigma}_{d_k}\rangle \langle \nu_{k}\rangle \sum_{n=1}^N \langle s_{k,n}\rangle {\boldsymbol{\Psi}}_n^T \langle \yv_{n,-k}\rangle, \\
\langle \yv_{n,-k}\rangle &= \yv_n - {\boldsymbol{\Psi}}_n\langle \boldsymbol{D} \rangle
\langle \boldsymbol{\Lambda}\rangle \langle \boldsymbol{s}_n\rangle +  {\boldsymbol{\Psi}}_n\langle{\boldsymbol{d}}_k\rangle \langle \nu_k\rangle  \langle s_{k,n}\rangle. \\
\langle {\bf \Sigma}_{d_k} \rangle&= \left(P\Imat_P + \langle\alpha_0 \rangle \langle \nu_k^2\rangle \sum_{n=1}^N \langle s_{k,n}^2\rangle {\boldsymbol{\Psi}}_n^T {\boldsymbol{\Psi}}_n\right)^{-1} , \\
\langle s_{k,n} \rangle &= \langle\alpha_{0}\rangle \langle\sigma^2_{s_{k,n}}\rangle \langle \nu_k \rangle  \langle{\boldsymbol{d}_k}\rangle^T {\boldsymbol{\Psi}}_n^T\langle{\boldsymbol{y}}_{n,-k}\rangle,\\ 
\langle s_{k,n}^2\rangle &= \langle s_{k,n} \rangle^2 + \langle \sigma^2_{s_{k,n}} \rangle,\\
\langle \sigma^2_{s_{k,n}} \rangle &= \left(\langle \tau_n\rangle \langle \alpha_{k,n} \rangle \langle \alpha_0\rangle + \langle\alpha_0\rangle \langle\nu_k^2\rangle  \sum^P_{p=1} \langle d_{p,k}^2\rangle w_{n,p} \right)^{-1}, \\
\langle d_{p,k}^2\rangle &= \langle d_{p,k}\rangle^2 + \langle \sigma_{d_{p,k}} \rangle,\quad
\langle\nu_k^2\rangle = \langle \nu_k \rangle^2 + \langle \sigma^2_{\nu_k}\rangle,\\
\langle \nu_k \rangle &= \langle \sigma_{\nu_k}^2 \rangle \langle \alpha_0\rangle \langle \dv_k \rangle^T \sum_{n=1}^N {\boldsymbol{\Psi}}_n^T \yv_{n,-k} \langle s_{k,n} \rangle,\\ 
\langle \sigma^2_{\nu_k} \rangle &= \left(\langle\eta_k \rangle + \langle\alpha_0\rangle \sum_{n=1}^N \langle s_{k,n}^2 \rangle \sum^P_{p=1} \langle d_{p,k}^2\rangle w_{n,p} \right)^{-1},
\end{align*}
\begin{eqnarray}
\langle \tilde{\eta}_j\rangle &=& \frac{e_0 + 0.5(K+1-j)}{f_0 + 0.5 \sum_{q=j}^K \langle \nu_q^2\rangle \frac{\langle\eta_q \rangle}{\tilde{\eta}_j}}, \\ 
\langle \tau_n\rangle &=& \frac{a_0 + 0.5K}{b_0 + 0.5 \sum_{k=1}^K \langle s_{k,n}^2\rangle \langle \alpha_{k,n} \rangle \langle \alpha_0\rangle}, \\
\langle \alpha_{k,n} \rangle &=& \sqrt{\frac{1}{\langle\Phi_{k,n}\rangle \langle s_{k,n}^2\rangle \langle \tau_n\rangle \langle \alpha_0\rangle}}, \\
\langle \Phi_{k,n} \rangle &=& \frac{\sqrt{\alpha_{k,n}^{-1}} K_{g_0}(\sqrt{2h_0 \alpha_{k,n}^{-1}})}{\sqrt{2 h_0} K_{g_0-1}(\sqrt{2h_0 \alpha_{k,n}^{-1}})}, \\
\langle \frac{1}{\Phi_{k,n}} \rangle &=& \frac{\sqrt{2 h_0}  K_{g_0+1}(\sqrt{2h_0 \alpha_{k,n}^{-1}})}{\sqrt{\alpha_{k,n}^{-1}}K_{g_0}(\sqrt{2h_0 \alpha_{k,n}^{-1}})}, \\
\langle \alpha_0 \rangle &=& \frac{c_0 + 0.5 \sum_n\|\boldsymbol{\Psi}_n\|_0 + 0.5 K N}{\begin{array}{c}
d_0 + 0.5 \sum_n \sum_k \langle s_{k,n}^2 \rangle \langle\tau_n\rangle \langle \alpha_{k,n}\rangle \\
+ 0.5 \sum_n \langle \|\yv_n - \boldsymbol{\Psi}_n\Dmat \boldsymbol{\Lambda s}_n\|_2^2 \rangle
\end{array}},
\end{eqnarray}
where 
\begin{eqnarray}
&&\langle \|\yv_n - \boldsymbol{\Psi}_n\Dmat {\bf \Lambda}  \boldsymbol{s}_n\|_2^2   \rangle \\
&&= {\mathbb E}[(\yv_n - \boldsymbol{\Psi}_n\Dmat {\bf \Lambda}\boldsymbol{s}_n)^T (\yv_n - \boldsymbol{\Psi}_n\Dmat {\bf \Lambda}\boldsymbol{s}_n)] \nonumber\\
&&= {\mathbb E}[\yv_n^T \yv_n] - 2 {\mathbb E} [\yv_n^T  \boldsymbol{\Psi}_n \Dmat {\bf \Lambda}  \boldsymbol{s}_n] +
{\mathbb E}[\boldsymbol{s}_n^T {\bf \Lambda} \Dmat^T \boldsymbol{\Psi}_n^T \boldsymbol{\Psi}_n \Dmat{\bf \Lambda} \boldsymbol{s}_n ] \nonumber\\
&&= \yv_n^T \yv_n - 2\yv_n^T \boldsymbol{\Psi}_n \langle \Dmat\rangle  \langle {\bf \Lambda}\rangle \langle \boldsymbol{s}_n \rangle  + {\mathbb E}[\rm Tr( \boldsymbol{\Psi}_n\Dmat  {\bf \Lambda} \boldsymbol{s}_n    \boldsymbol{s}_n^T  {\bf \Lambda}\Dmat^T \boldsymbol{\Psi}_n^T)]  \nonumber\\
&&= \yv_n^T \yv_n - 2\yv_n^T  \langle \Dmat\rangle \langle {\bf \Lambda}\rangle \langle \boldsymbol{s}_n \rangle  + {\rm Tr} ([ \langle \boldsymbol{s}_n \boldsymbol{s}_n^T\rangle  \langle{\bf \Lambda}\Dmat^T \boldsymbol{\Psi}_n^T \boldsymbol{\Psi}_n \Dmat {\bf \Lambda} \rangle]), \nonumber
\end{eqnarray}
with ${\rm Tr (\cdot)}$ denoting the trace of the matrix inside $(\hspace{0.3mm})$.
\begin{eqnarray}
\langle \boldsymbol{s}_n \boldsymbol{s}_n^T\rangle &=& \langle \boldsymbol{s}_n \rangle
\langle \boldsymbol{s}_n^T \rangle  + {\bf \Sigma}_{\boldsymbol{s}_n},\\
{\bf \Sigma}_{\boldsymbol{s}_n} &=& {\rm diag} [\sigma^2_{s_{1,n}},\dots, \sigma^2_{s_{K,n}} ],\\
\langle{\bf \Lambda}\Dmat^T \boldsymbol{\Psi}_n^T \boldsymbol{\Psi}_n \Dmat {\bf \Lambda} \rangle &=& \langle {\boldsymbol{\nu}} {\boldsymbol{\nu}}^T\rangle \odot \langle \Dmat^T \boldsymbol{\Psi}_n^T \boldsymbol{\Psi}_n \Dmat \rangle,\\
 {\boldsymbol{\nu}} &=& [\nu_1,\dots, \nu_K]^T,\\
 \langle {\boldsymbol{\nu}} {\boldsymbol{\nu}}^T\rangle &=&  \langle {\boldsymbol{\nu}} \rangle \langle {\boldsymbol{\nu}}^T\rangle +  {\rm diag}[\sigma^2_{\nu_1},\dots, \sigma^2_{\nu_K}],\\
\langle\Dmat^T \boldsymbol{\Psi}_n^T \boldsymbol{\Psi}_n \Dmat \rangle &=& \langle \boldsymbol{\Psi}_n\Dmat \rangle^T \langle \boldsymbol{\Psi}_n\Dmat \rangle \nonumber\\
&&\hspace{-20mm}+ {\rm diag}\left(\sum_{p}\sigma^2_{d_{p,1}} \Psi_{n,p}, \dots, \sum_{p}\sigma^2_{d_{p,K}} \Psi_{n,p}\right).
\end{eqnarray}

\section{More Results}
\label{Sec:Result}

\subsection{Benchmark Images for Inpainting and Denoising}
\begin{figure*}[ht!]
  \centering
  \includegraphics[scale = 0.45]{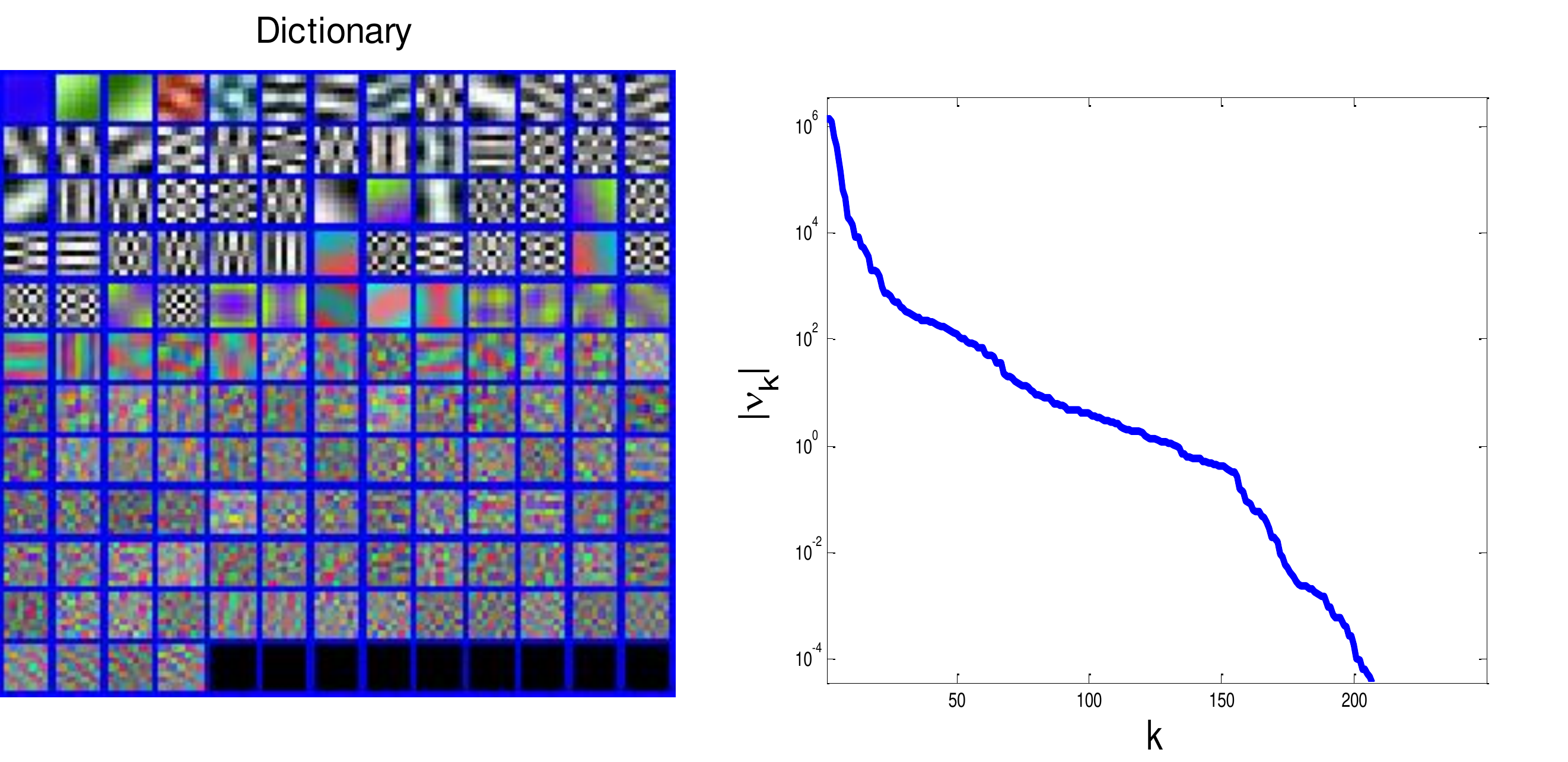}\\
\caption{\small{Left: dictionary learned from the noisy image ``kangaroo" with $\sigma=5$ and patch size $7\times 7 \times 3$. Though we set $K=256$, only 147 of these atoms are significant. Right: inferred $|\nu_k|$ (sorted by absolute values from large to small).}}
  \label{Fig:Dict}
\end{figure*}

\begin{figure*}[ht!]
  \centering
  \includegraphics[scale = 0.8]{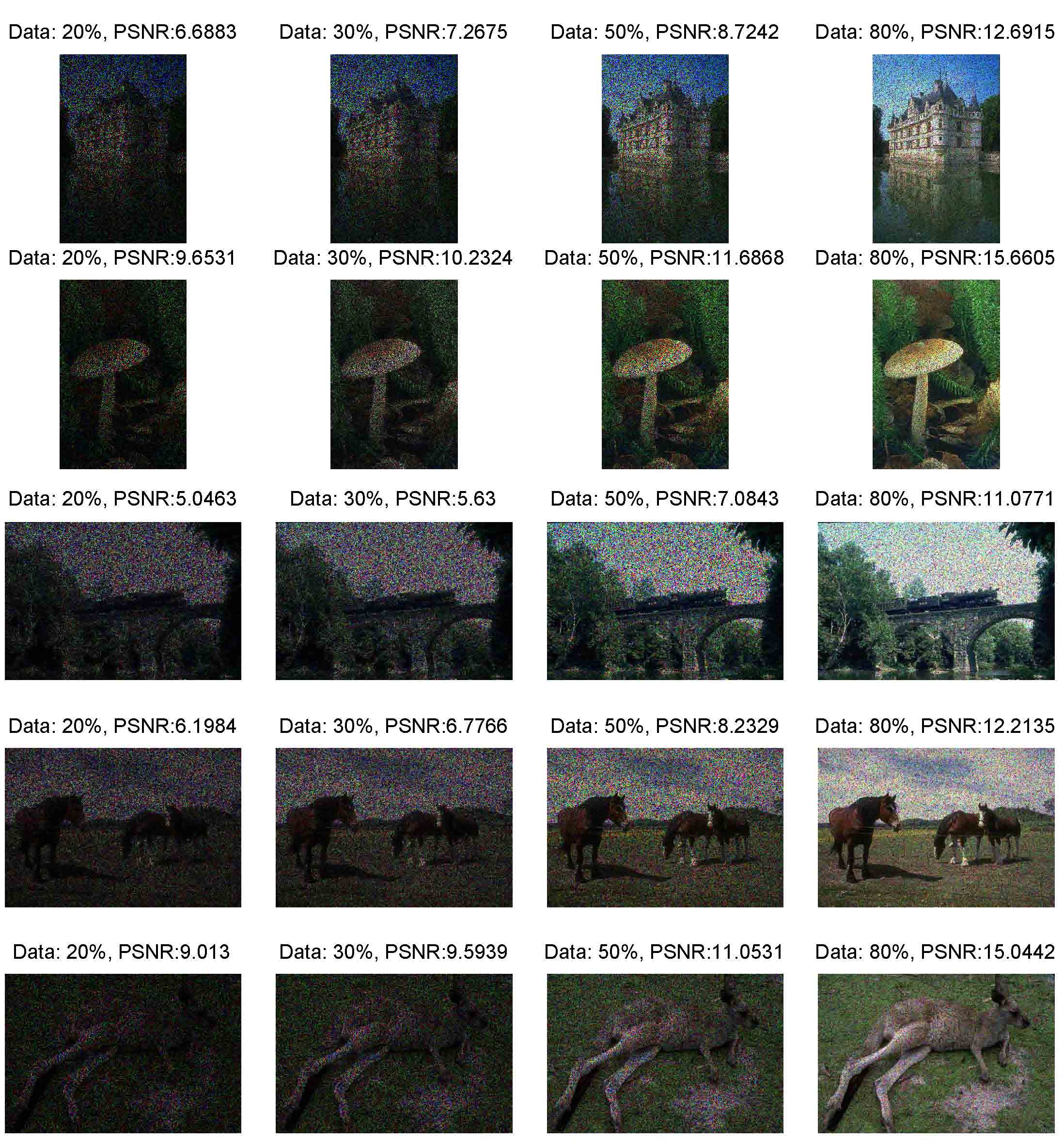}\\
\caption{\small{Corrupted images with missing values for inpainting.  Each row shows one image, each column shows one data ratio (observed pixel precentage).}}
  \label{Fig:Mea_inpainting}
\end{figure*}

\begin{figure*}[ht!]
  \centering
  \includegraphics[scale = 0.8]{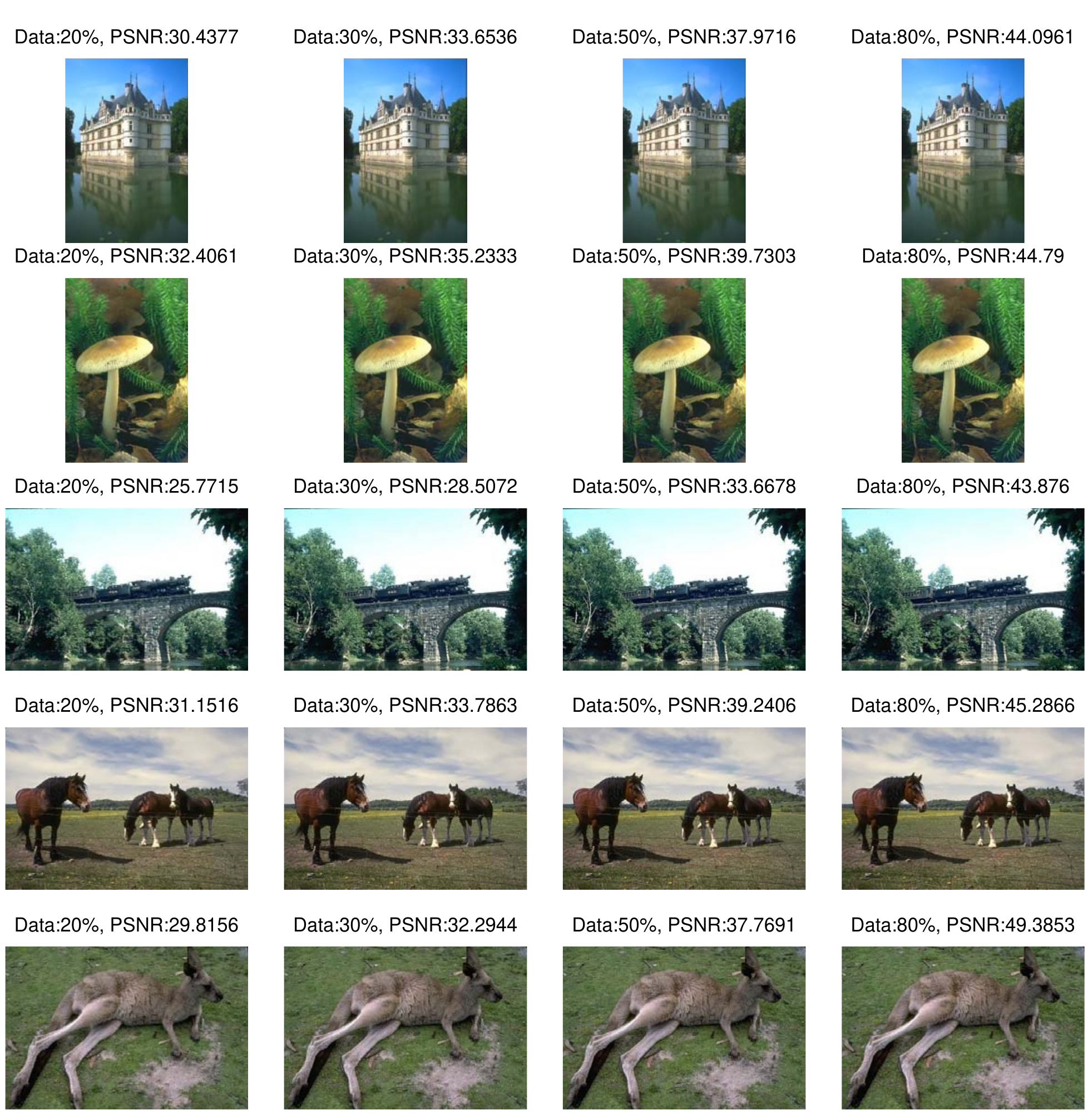}\\
\caption{\small{Restored images of inpainting; each row shows one image; each column shows one data ratio.}}
  \label{Fig:recon_inpainting}
\end{figure*}

\begin{figure*}[ht!]
  \centering
  \includegraphics[scale = 0.84]{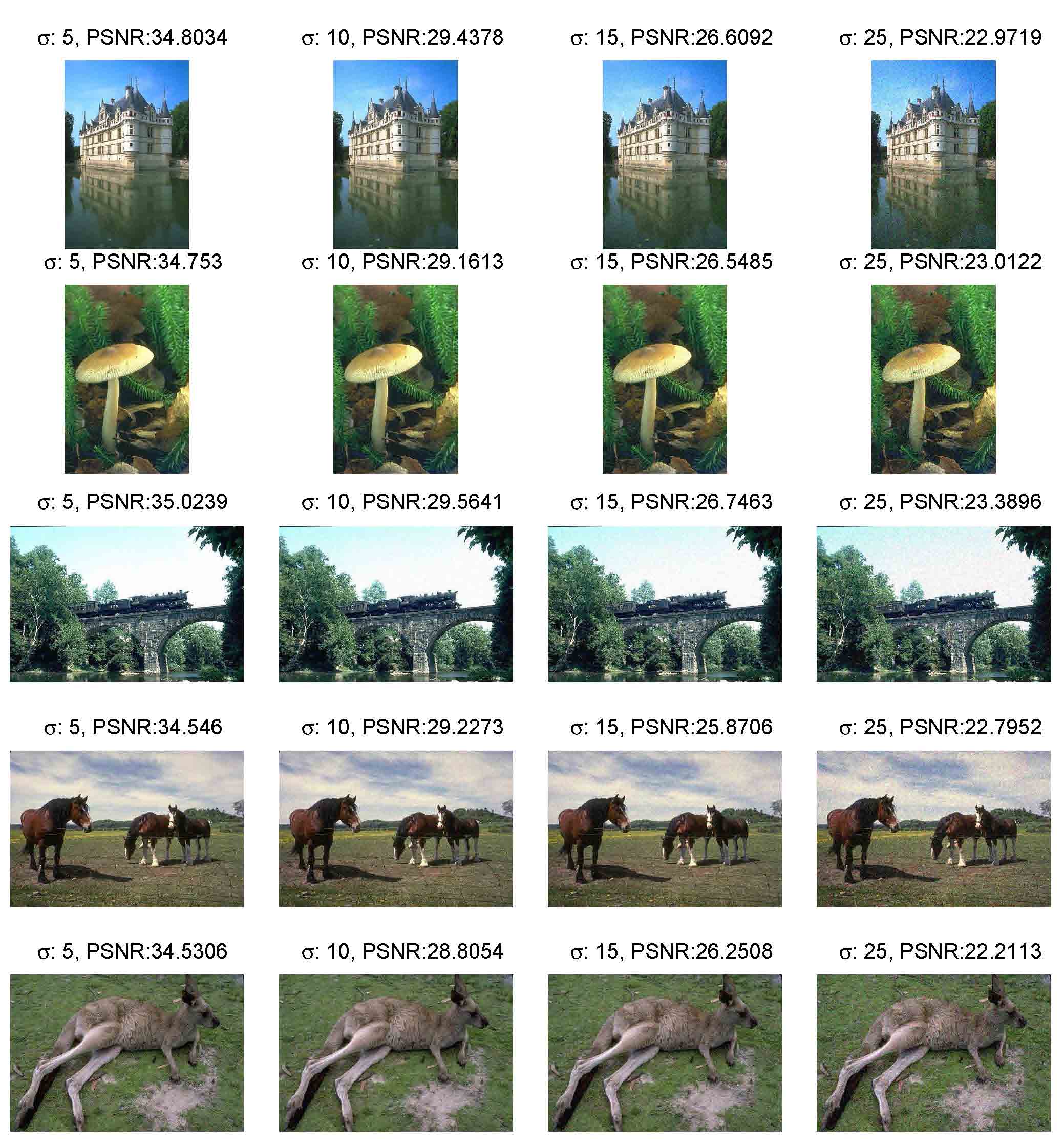}\\
\caption{\small{Noisy images for denoisng. Each row shows one image; each column shows one noise level, defined by standard deviation $\sigma$ of the additive zero-mean Gaussian noise.}}
  \label{Fig:noise}
\end{figure*}

\begin{figure*}[ht!]
  \centering
  \includegraphics[scale = 0.8]{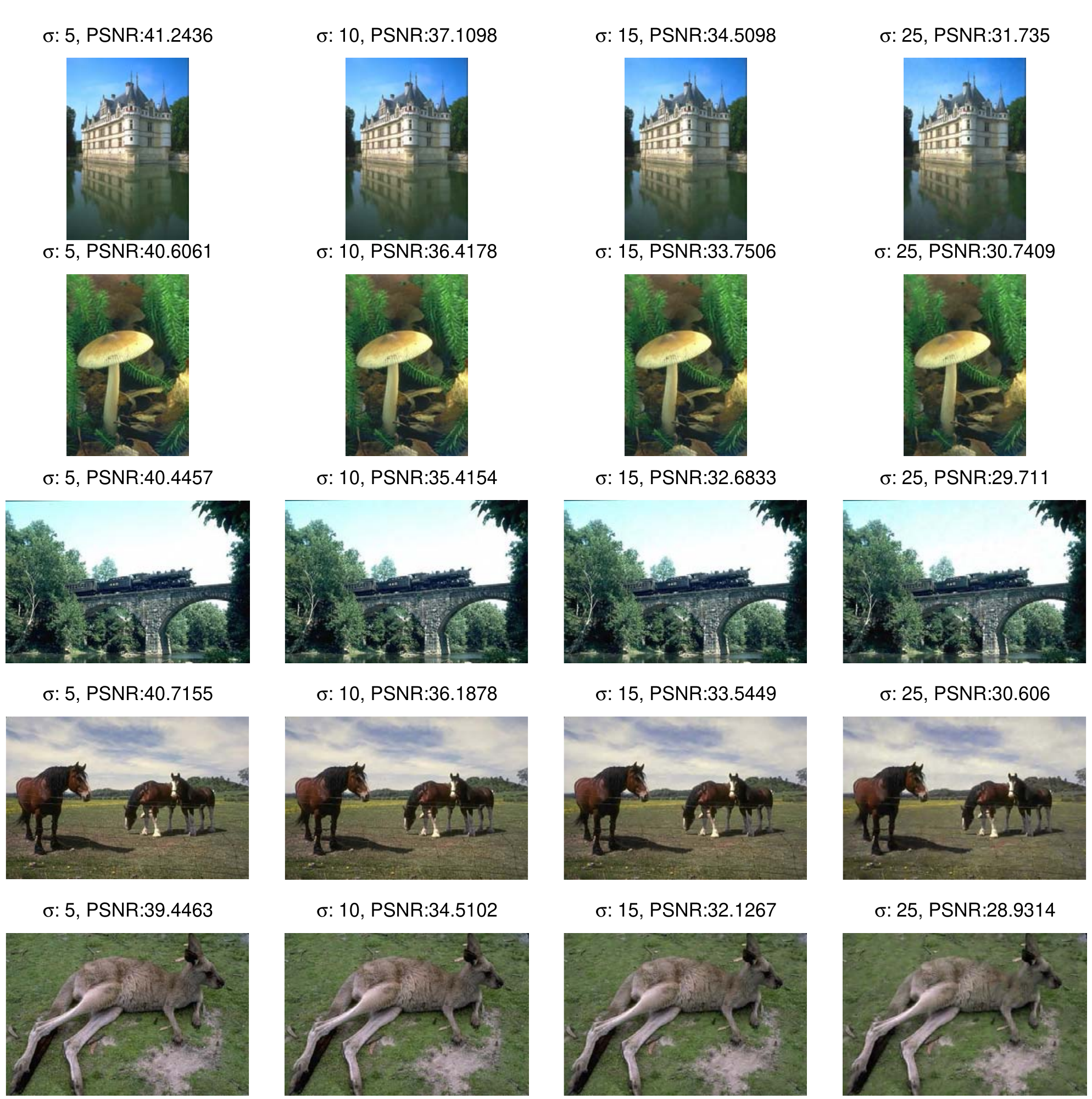}\\
\caption{\small{Denoised images.  Each row shows one image; each column shows one standard deviation $\sigma$.}}
  \label{Fig:denoise}
\end{figure*}

To demonstrate the general applications of our proposed dictionary learning model, we show the denoising and inpainting results from benchmark color images (8-bits) tested in \cite{Mairal07TIP,Zhou12TIP}.  For inpainting, we show PSNRs of the restored images at various observed data ratios ($20\%$ means $80\%$ pixels are missing) in Table~\ref{Table:InpaintingPSNR}.
Noisy images corrupted with zero-mean Gaussian noise with different standard deviations $\sigma$ are restored with PSNRs shown in Table~\ref{Table:DenoisngPSNR}.
Our proposed algorithm consistently provides the highest PSNR.
Inpainting results for corrupted images (Figure~\ref{Fig:Mea_inpainting}) are presented in Figure~\ref{Fig:recon_inpainting} (PSNRs shown in Table~\ref{Table:InpaintingPSNR}); 
denoising results for noisy images (corrupted with zero-mean Gaussian noise of various standard deviations, Figure~\ref{Fig:noise}) are presented in Figure~\ref{Fig:denoise} (PSNRs shown in Table~\ref{Table:DenoisngPSNR}).
Figure~\ref{Fig:Dict} shows an example of a dictionary and $\nu_k$ learned from a noisy image using the proposed dictionary learning model (Section~\ref{Sec:Model}).
Importantly, few iterations of our VB inference are required to obtain good results; 20 iterations are used for denoising and 100 iterations are used for inpainting ($\sim$1 second per iteration with an image size $256\times 256\times 3$ on an i5 CPU with non-optimized MATLAB code).

\begin{table}[ht!]
\caption{\small{Inpainting results of color images at various observed pixel percentages.  We compare BPFA~\cite{Zhou12TIP} (top row in each cell) with the proposed algorithm (bottom row in each cell). 8-bit RGB images are used.}}
\centering
\tiny
\begin{tabular}{||c||c|c|c|c|c||}
\hline data ratio &  Castle& Mushroom& Train & Horses& Kangaroo \\
\hline\hline 20\% &
$\begin{array}{c} 29.12\\ {\bf 30.44} \end{array}$    &
$\begin{array}{c} 31.56 \\ {\bf 32.41 } \end{array}$  &
$\begin{array}{c} 24.59\\ {\bf  25.77}  \end{array}$ &
$\begin{array}{c} 29.99\\  {\bf  31.15}   \end{array}$ &
$\begin{array}{c} 29.59\\  {\bf 29.82}   \end{array}$ \\
\hline
30\% &$\begin{array}{c} 32.02\\  {\bf 33.65} \end{array}$    &
$\begin{array}{c} 34.63 \\ {\bf 35.23} \end{array}$  &
$\begin{array}{c} 27.00 \\ {\bf 28.51}   \end{array}$ &
$\begin{array}{c} 32.52 \\ {\bf  33.78}   \end{array}$ &
$\begin{array}{c} 32.21\\  {\bf  32.29}   \end{array}$ \\
\hline
50\% &$\begin{array}{c} 36.45 \\ {\bf 37.97} \end{array}$    &
$\begin{array}{c} 38.88\\ {\bf 39.73}  \end{array}$  &
$\begin{array}{c} 32.00 \\ {\bf 33.67 }   \end{array}$ &
$\begin{array}{c} 37.27\\{\bf  39.24}   \end{array}$ &
$\begin{array}{c} 37.34 \\ {\bf  37.77}   \end{array}$ \\
\hline
80\% &$\begin{array}{c} 41.51\\ {\bf 44.10}  \end{array}$
& $\begin{array}{c} 42.56\\  {\bf 44.79 }  \end{array}$  &
$\begin{array}{c} 40.73 \\ {\bf  43.88}   \end{array}$ &
$\begin{array}{c} 41.97 \\   {\bf  45.29}  \end{array}$ &
$\begin{array}{c}  42.74\\ {\bf  49.39}   \end{array}$ \\
\hline\hline
\end{tabular}
\label{Table:InpaintingPSNR}
\end{table}

\begin{table}[ht!]
\caption{\small{Denoising results of color images at various noise levels; $\sigma$ denotes the noise standard deviations.  We compare KSVD~\cite{Mairal07TIP} (top row in each cell), with BPFA~\cite{Zhou12TIP} (middle row in each cell) and the proposed algorithm (bottom row in each cell).}}
\centering
\begin{tabular}{||c||c|c|c|c|c||}
\hline $\sigma$ &  Castle& Mushroom& Train & Horses& Kangaroo \\
\hline\hline 5 &$\begin{array}{c} 40.37\\ 40.34\\ {\bf 41.24} \end{array}$    &
$\begin{array}{c} 39.93\\ 39.73 \\ {\bf 40.60} \end{array}$  &
$\begin{array}{c} 39.76\\ 39.38\\ {\bf 40.45}  \end{array}$ &
$\begin{array}{c} 40.09 \\ 39.96\\  {\bf 40.72}   \end{array}$ &
$\begin{array}{c} 39.00\\ 39.00\\  {\bf 39.25}   \end{array}$ \\
\hline
10 &$\begin{array}{c} 36.24 \\ 36.28\\  {\bf 37.11} \end{array}$    &
$\begin{array}{c} 35.60\\ 35.70 \\ {\bf 36.42} \end{array}$  &
$\begin{array}{c} 34.72 \\ 34.48 \\ {\bf 35.42}   \end{array}$ &
$\begin{array}{c} 35.43 \\ 35.48 \\ {\bf 36.19}   \end{array}$ &
$\begin{array}{c} 34.06 \\ 34.21\\  {\bf 34.26}   \end{array}$ \\
\hline
15 &$\begin{array}{c} 33.98\\ 34.04 \\ {\bf 34.51} \end{array}$    &
$\begin{array}{c} 33.18\\ 33.41\\ {\bf 33.75}  \end{array}$  &
$\begin{array}{c} 31.70 \\ 31.63 \\ {\bf 32.68}   \end{array}$ &
$\begin{array}{c} 32.76 \\ 32.98\\{\bf 33.55}   \end{array}$ &
$\begin{array}{c} 31.30\\ 31.68 \\ {\bf 32.13}   \end{array}$ \\
\hline
25 &$\begin{array}{c} 31.19\\ 31.24\\ {\bf 31.74}  \end{array}$
& $\begin{array}{c} 30.26\\ 30.62 \\ {\bf 30.74}  \end{array}$  &
$\begin{array}{c} 28.16 \\ 28.28 \\ {\bf 29.71}   \end{array}$ &
$\begin{array}{c} 29.81 \\ 30.11 \\ {\bf 30.61}   \end{array}$ &
$\begin{array}{c}  28.39\\ 28.86\\ {\bf 28.93}   \end{array}$ \\
\hline\hline
\end{tabular}
\label{Table:DenoisngPSNR}
\end{table}

\end{document}